%% file: main.tex
\documentclass[prodmode]{acmlarge}

\acmVolume{0}
\acmNumber{0}
\acmArticle{0}
\articleSeq{0}
\acmYear{2017}
\acmMonth{0}

\doi{0000001.0000001}

\issn{1234-56789}

\usepackage[ruled]{algorithm2e}
\SetAlFnt{\algofont}
\SetAlCapFnt{\algofont}
\SetAlCapNameFnt{\algofont}
\SetAlCapHSkip{0pt}
\IncMargin{-\parindent}

\title{Ensembles of Deep LSTM Learners for Activity Recognition using Wearables}
\author{Yu Guan 
\affil{Open Lab, Newcastle University, Newcastle upon Tyne, UK}
Thomas Pl{\"o}tz
\affil{{\color{black}School of Interactive Computing, Georgia Institute of Technology, Atlanta, USA}}
}

\begin{abstract}
Recently, deep learning (DL) methods have been introduced very successfully into human activity recognition (HAR) scenarios in ubiquitous and wearable computing.
Especially the prospect of overcoming the need for manual feature design combined with superior classification capabilities render deep neural networks very attractive for real-life HAR applications.
Even though DL-based approaches now outperform the state-of-the-art in a number of recognition tasks, still substantial challenges remain.
Most prominently, issues with real-life datasets, typically including imbalanced datasets and problematic data quality, still limit the effectiveness of activity recognition using wearables.
In this paper we tackle such challenges through Ensembles of deep Long Short Term Memory (LSTM) networks.
{\color{black}LSTM networks currently represent the state-of-the-art with superior classification performance on relevant HAR benchmark datasets.}
We have developed modified training procedures for LSTM networks and combine sets of diverse LSTM learners into classifier collectives.
We demonstrate that Ensembles of deep LSTM learners outperform individual LSTM networks {\color{black}and thus push the state-of-the-art in human activity recognition using wearables}.
Through an extensive experimental evaluation on three standard benchmarks (Opportunity, PAMAP2, Skoda) we demonstrate the excellent recognition capabilities of our approach and its potential for real-life applications of human activity recognition.
\end{abstract}


\acmformat{Yu Guan, and Thomas Pl{\"o}tz. 2017. Ensembles of Deep LSTM Learners for Activity Recognition using Wearables}

\usepackage[textsize=small]{todonotes}
\usepackage{amsfonts}
\usepackage{amsmath}

\usepackage{graphicx}
\usepackage{epsfig}

\usepackage{here}

\usepackage[export]{adjustbox}

\usepackage[caption=false]{subfig}

\begin{document}

\begin{bottomstuff}
Authors' addresses: Y. Guan, 89 Sandyford Road, Newcastle University, Newcastle upon Tyne, NE1 8HW, UK; email: yu.guan@newcastle.ac.uk;
T.\ Pl{\"o}tz, Technology Square Research Building, Georgia Institute of Technology, 85 Fifth Street NW, Atlanta, GA 30308; email: thomas.ploetz@gatech.edu
\end{bottomstuff}

\maketitle

\input{introduction}

\input{related}

\input{system}
\input{experiments}

\input{discussion}

\input{appendix}

\section*{Acknowledgements}
{\color{black}
We would like to thank the School of Computing Science at Newcastle University for providing the computing facilities (GPU NVIDIA Tesla K40), and the anonymous reviewers for their constructive comments that helped improving the manuscript of this paper.}


\bibliographystyle{ACM-Reference-Format-Journals}
\bibliography{main}

\end{document}

%% file: introduction.tex

\section{Introduction}
\label{sec:introduction}
Activity recognition has been a core concern of ubiquitous and wearable computing ever since Weiser formulated his vision of the third generation of computing \cite{Weiser1991}.
Related work now covers an enormous variety of innovative sensing and analysis approaches that target a plethora of application areas, including but not limited to novel interaction techniques \cite{Kim2012}, situated support in smart environments \cite{Hoey:2011up}, automated health \cite{Ploetz:2012vm} and wellbeing assessments \cite{Kranz:2012us}, health care automation \cite{Avci:2010tz}, sports tracking and coaching \cite{Ladha:2013wb}  to name but a few. 
In fact, the field has advanced so much that it is now already focusing on application areas that go beyond activity recognition, such as automated skill or quality assessments (e.g., \cite{Khan2015}).
Still, substantial challenges for human activity recognition in ubiquitous and wearable computing remain and they are manifold.
First and foremost, data recorded using wearable and pervasive sensing modalities\footnote{Without limiting our scope, in this paper we primarily refer to data recorded using inertial measurement units (IMUs).} are inherently noisy, often containing missing or even erroneous readings due to sensor malfunction.
Furthermore, ground truth annotation for mobile application scenarios is often hard to obtain if not impossible to collect.
Accurate labels are, however, required for, e.g., supervised learning methods but also for validation in general.
This is in stark contrast to the almost effortless collection of (unlabeled) sensor data through mobile devices.
Also, real-world HAR datasets tend to be imbalanced with regards to a biased class distribution of the underlying sensor data, which causes problems at both training and inference time.

Encouraged by the tremendous success of deep learning (DL) methods, recently the activity recognition research community has started adopting more complex modelling and inference procedures.
These deep learning methods have the {\color{black}proven} potential to significantly push the state-of-the-art in human activity recognition.
Most importantly they allow to overcome the ever so crucial need for feature engineering, that is, the manual definition of feature extraction procedures, which is often an erroneous or --at the very least-- poorly generalisable endeavour.
The biggest advantage of contemporary deep learning methods is their ability to simultaneously learn both proper data representations and classifiers.
Early feature learning methods in Ubicomp utilised restricted Boltzmann machines (RBM), that is, generative deep learners for deriving task agnostic feature representations \cite{Ploetz2011-FLF,Hammerla2015}.
Recently, more sophisticated models have successfully been used for challenging HAR tasks, e.g., convolutional neural networks (CNNs) \cite{Ordonez2016}. 

A popular way of using deep learning methods for activity recognition purposes is by modifying the "standard" processing workflow as it has been widely adopted by the community \cite{Bulling2014}. 
This workflow comprises sliding window based analysis of isolated frames, that is, consecutive (typically overlapping) portions of sensor readings.
Instead of manually defining feature representations for these frames, the raw data are presented to the DL models that automatically extract rich  representations of the input data.
The use of multiple hidden layers of typically fully connected computing nodes then allows for complex function approximation that leads to very powerful classifiers. 
As it stands, deep learning based systems outperform alternatives on standard, challenging activity recognition benchmarks such as Opportunity \cite{Chavarriaga2013a}.

Whilst the predominant modelling approach for deep learning based HAR is based on (variants of) CNN, recently also sequential modelling methods have been employed very successfully, e.g., \cite{Ordonez2016,Hammerla2016}.
Specifically, so-called Long Short Term Memory (LSTM) networks \cite{Hochreiter1997} have been used, that is, recurrent neural networks with principally infinite memory for every computing node.
LSTM models, like any other recurrent neural network, are predestined for the analysis of sequential data such as streams of sensor data.
Processing nodes, i.e., neurons, also referred to as cells, not only take spatial but also temporal context into account when determining their activation.
In addition to connectivity across the network, these models implement feedback loops where outputs of neurons directly serve as their input as well.
Through adding specific elements to the individual cells, effective memory functionalities can be realised \cite{Hochreiter1997}.
{\color{black}Most recent works in the ubiquitous and wearable computing arena have substantially pushed LSTM-based recognition systems, rendering them state-of-the-art by outperforming other approaches on relevant, challenging baseline tasks. As such, deep learning in general and LSTM-based modelling approaches in particular serve as reference for the work presented in this paper.}

Even though deep learning based approaches to activity recognition widely outperform the state-of-the-art using non-DL methods, some of the aforementioned challenges for HAR in ubiquitous and wearable computing remain.
Specifically, the problematic data situation --noisy / erroneous data, heavily imbalanced data distributions etc.-- remains largely unaddressed. 
Furthermore, it is rather surprising to see that models that have been specifically designed for the analysis of sequential data are widely employed --during inference-- in the traditional sliding-window, i.e., frame-based manner.

In this paper we present a framework for deep learning based activity recognition using {\color{black}wearable sensing data}. 
{\color{black}We substantially extend our previous work where we have alleviated the sliding window paradigm of sensor data processing \cite{Hammerla2016}. We employ deep, recurrent LSTM networks for sample-wise prediction. For the first time, we combine multiple LSTM learners into ensemble classifiers that enable substantially improved robustness for challenging, realistic HAR applications.}
Directly tackling practical issues of real-world sensing scenarios we develop a novel training method for LSTM models that leads to the derivation of more robust recognisers.
This method acknowledges the fact that noisy, possibly ambiguous or even erroneous data --as they are typical for HAR scenarios in Ubicomp-- negatively influence the performance of recognition systems.
{\color{black}It is worth mentioning that in real-world HAR deployments one cannot easily identify and subsequently ignore the NULL class, that is data that do not belong to any of the activities of interest but rather to the background. 
The larger context of our work aims at 'open-ended' activity recognition, that is at deriving and adapting recognition systems without assumptions and limitations that would hinder actual, 'in-the-wild' deployments.
As such pre-filtering based on prior domain knowledge does not represent a reasonable option and we rather aim for a generalised approach, as presented in this paper.}

Our training method assumes that a certain percentage of sensor data is 'problematic' and thus only utilises a subset of the data in a training iteration.
Given that such data issues are normally difficult to detect and specify \textit{a-priori},  our subset determination method employs a probabilistic selection approach, which effectively resembles the general concept of Bagging \cite{Breiman1996-BP}.
Through reiterating the selection procedure we are able to build ensemble classifiers for activity recognition.
To the best of our knowledge this paper is the first that uses Ensembles of LSTM learners for an activity recognition scenario.
We evaluate our method on three standard benchmark datasets, namely the Opportunity challenge \cite{Chavarriaga2013a}, PAMAP2 \cite{Reiss2012}, and the Skoda dataset \cite{Stiefmeier2008-WAT}.
By outperforming the state-of-the-art our results indicate increased robustness of our Ensembles of deep LSTM learners with regards to typical challenges of human activity recognition in ubiquitous and wearable computing scenarios.
We provide a detailed analysis of our models' performance, which allows us to draw conclusions for related real-world applications and future scenarios.

The remainder of this paper is structured as follows. 
In Sec.\ \ref{sec:related} we provide a 
{\color{black} primer on the relevant background in deep learning for human activity recognition, as well as in Ensemble learning. 
This overview serves as reference for the technical developments described later in the paper (and beyond for the wider Ubicomp audience).}
Subsequently, we explain in detail our framework for activity recognition using Ensembles of deep LSTM learners.
Sec.\ \ref{sec:experiments} presents our experimental evaluation based on the aforementioned three benchmark datasets and discusses results with a view on real-world applications.
Sec. \ref{sec:discussion} summarises our work and we discuss {\color{black}in detail} practical implications as well as future avenues for extensions and adaptations.

%% file: related.tex

\section{Background}
\label{sec:related}

We aim for developing methods for robust human activity recognition (HAR) using ubiquitous and wearable sensing modalities.
Most prominently but without limiting ourselves we target inertial measurement units (IMUs).
HAR has a long standing history in the wider ubiquitous and wearable computing community.
Over the years a multitude of methods have been developed that facilitate an astonishing variety of applications.
HAR has become one of the pillars of the third generation of computing \cite{Schmidt1999-TIM} and it is more than likely that future developments will depend on robust and reliable activity recognition from sensor data as well \cite{Abowd:2016vv}.

Aiming for conciseness we refrain here from reiterating summaries of the general state-of-the-art in activity recognition.
We argue that HAR can now safely be considered common knowledge and thus refer to seminal papers and tutorials such as \cite{Bulling2014,Chen:2012gt,Avci2010-ARU}.
More importantly, in what follows we will cover the specific algorithmic background for this paper, which spans the following two main subject areas:
\textit{i)} Deep learning for HAR in ubiquitous and wearable computing; and
\textit{ii)} Ensembles of classifiers for robust recognition in challenging settings.



%

\subsection{Deep Learning for Human Activity Recognition}
\label{sec:related:DLHAR}
In line with its massive success and popularity in many application domains, deep learning \cite{LeCun:2015dt} is also about to revolutionise human activity recognition methods in the field of ubiquitous and wearable computing.
The attraction of deep learning not only stems from the fact that the complex models come with great capacity with regards to classification tasks, they also substantially alleviate the need for feature engineering as they learn very effective, rich representations of input data.

Early deep learning applications in ubiquitous and wearable computing primarily targeted this representation learning aspect. 
In \cite{Ploetz2011-FLF} deep belief networks, specifically generative Restricted Boltzmann Machines (RBMs) were employed for feature learning. 
Based on this, subsequent works explored the effectiveness of pre-trained, fully-connected RBM networks, for example, for automated assessments of disease states in Parkinson's patients \cite{Hammerla2015}, in combination with more traditional sequence models (HMMs) \cite{Zhang:2015hc,Alsheikh:2015vz}, and very successfully for auditory scene analysis in Ubicomp applications \cite{lane2015deepear}.

Most popular in the field of human activity recognition using wearables are two variants of deep learning methods:
\textit{i)} Deep Convolutional Neural Networks (CNNs); and
\textit{ii)} Recurrent Deep Neural Networks, such as Long Short Term Memory (LSTM) networks.
Both types and their use in HAR are summarised in what follows.

\subsubsection{Convolutional Neural Networks for HAR}
\label{sec:related:CNN}
Arguably the most widely used deep learning approach in the ubiquitous computing field in general and in human activity recognition using wearables in particular employ CNNs.
CNNs typically contain multiple hidden layers that implement convolutional filters that extract abstract representations of input data.
Combined with pooling and/ or subsampling layers, and fully connected special layers, CNNs are able to learn hierarchical data representations and classifiers that lead to extremely effective analysis systems.
A multitude of applications are based on CNNs, including but not limited to \cite{Zeng:2014tk,Ronaoo:2015wm,Yang:2015ts,Rad:2015vz}.


Recently, sophisticated model optimisation techniques have been introduced that actually allow for the implementation of deep CNNs in resource constrained scenarios, most prominently for real-time sensor data analysis on smartphones and even smart watches \cite{Bhattacharya:wd}.

\subsubsection{Long Short Term Memory Models for HAR}
\label{sec:related:LSTM}
The de-facto standard workflow for activity recognition in ubiquitous and wearable computing \cite{Bulling2014} treats individual frames of sensor data as statistically independent, that is, isolated portions of data are converted into feature vectors that are then presented to a classifier without further temporal context.
{\color{black}However, ignoring the temporal context beyond frame boundaries during modelling may limit the recognition performance for more challenging tasks.}
Instead, approaches that specifically incorporate temporal dependencies of sensor data streams seem more appropriate for human activity recognition.
In response to this, recurrent deep learning methods have now gained popularity in the field.
Most prominently models based on so-called LSTM units \cite{Hochreiter1997} have been used very successfully.
In \cite{Ordonez2016} deep recurrent neural networks have been used for activity recognition on the Opportunity benchmark dataset.
The LSTM model was combined with a number of preceding CNN layers in a deep network that learned rich, abstract sensor representations and very effectively could cope with the non-trivial recognition task.
Through large scale experimentation in \cite{Hammerla2016} appropriate training procedures have been analysed for a number of deep learning approaches to HAR including deep LSTM networks.
{\color{black}In all of previous work, including our own \cite{Hammerla2016}, single LSTM models have been used and standard training procedures have been employed for parameter estimation. The majority of existing methods is based on (variants of) sliding-window procedures for frame extraction. The focus of this paper is on capturing diversity of the data during training and to incorporate diverse models into Ensemble classifiers. As such this paper is the first that pushes beyond optimisation of individual models (as demonstrated, for example, in the meta-study in \cite{Hammerla2016}), which leads to significantly improved recognition performance and robustness.}

\paragraph*{Theoretical Background for LSTMs}

\begin{figure}[tp]
	\centering
	\includegraphics[height=7cm]{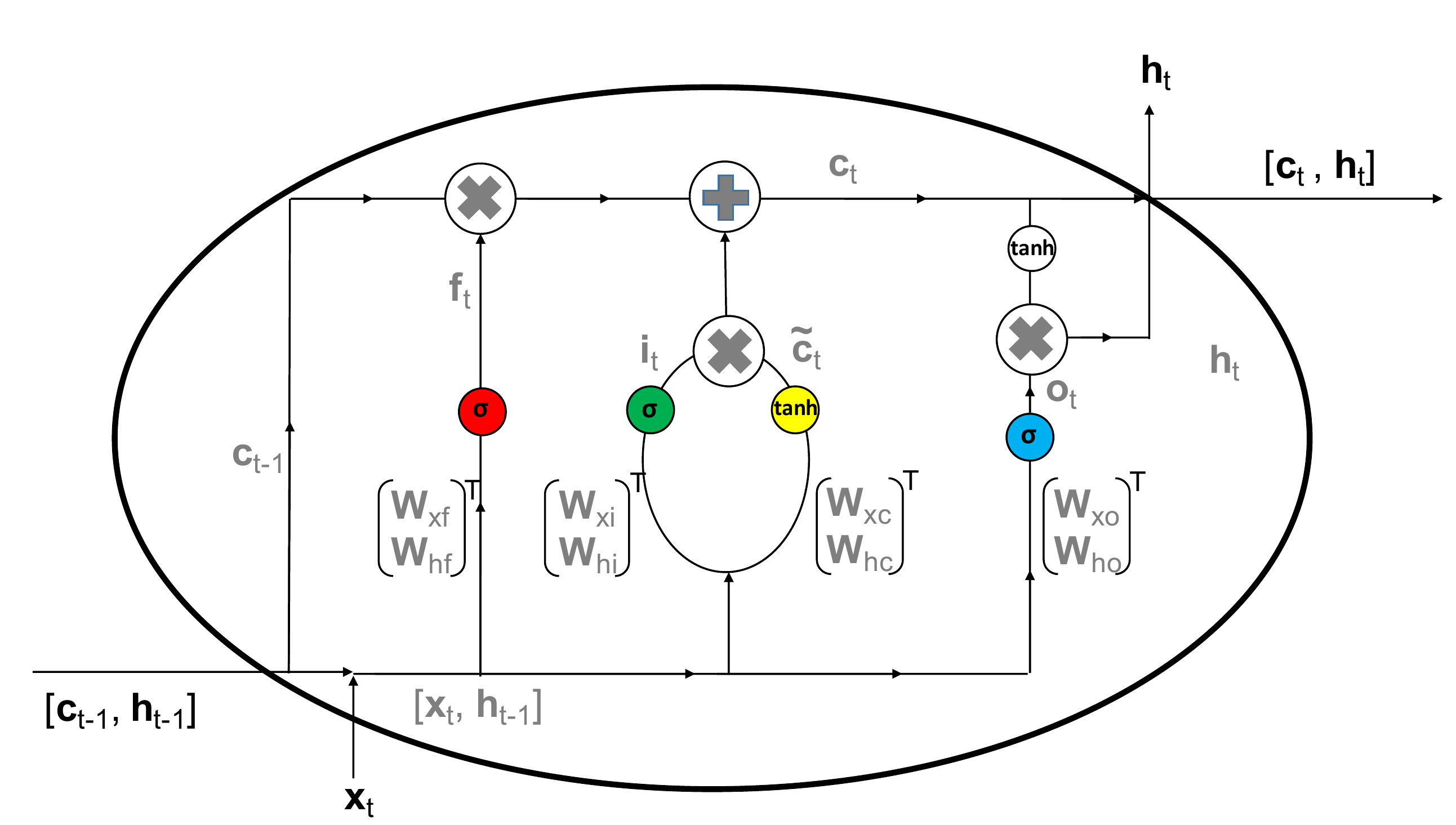}
	\caption{LSTM unit according to \protect\cite{Hochreiter1997}. Red, green, blue, and yellow highlighted parts refer to forget, input, and output gate, as well as the cell. Bias vectors are not shown here for better clarity  (see text for description).}
	\label{fig:LSTM_unit}	
\end{figure}

Given that LSTM networks form the algorithmic foundations for the work presented in this paper, we will now briefly summarise their theoretical background.
Fig.\ \ref{fig:LSTM_unit} schematically illustrates the anatomy of LSTM units as originally introduced in \cite{Hochreiter1997}.
In order to address the problem of vanishing/ exploding gradients as it is typical for large scale vanilla (recurrent) networks\footnote{Vanilla networks are those that are trained using the traditional backpropagation algorithm \cite{McClelland:1986we}.} based on $\tanh$ activation functions, specific mechanisms were introduced for every individual hidden unit of an LSTM network.
These include forget, input, cell, and output gates, as well as a specific internal structure of individual hidden units -- all of which facilitating effective error back-propagation even for very complex, that is deep model architectures and over prolonged periods of time, e.g., hundreds of time steps.

For an exemplary one-layer network, the LSTM units take the input signal $\mathbf{x}_t$, the hidden state  $\mathbf{h}_{t-1}$ and the cell state $\mathbf{c}_{t-1}$ from previous LSTM units.
The output forwarded consists of the current hidden state $\mathbf{h}_t$ as well as the cell state $\mathbf{c}_t$.
The whole LSTM network has four sets of specific gate parameters and one set of network output layer parameters, which are defined as follows:

\begin{eqnarray}
 	\text{forget gate  parameters:} 	\quad\quad &
		\begin{bmatrix} \mathbf{W_{xf}} \\ \mathbf{W_{hf}} \end{bmatrix}  &
		\in \mathbb{R}^{(D+H)\times H},\, \mathbf{b_f}\in \mathbb{R}^H  \quad\quad\quad\quad\quad\quad	\\
	\text{input gate parameters:}  	\quad\quad &
		\begin{bmatrix} \mathbf{W_{xi}} \\ \mathbf{W_{hi}} \end{bmatrix}  &
		\in \mathbb{R}^{(D+H)\times H},\, \mathbf{b_i}\in \mathbb{R}^H	\\
	\text{cell parameters:} 		\quad\quad &
		\begin{bmatrix} \mathbf{W_{xc}} \\ \mathbf{W_{hc}} \end{bmatrix}  &
		\in \mathbb{R}^{(D+H)\times H},\, \mathbf{b_c}\in \mathbb{R}^H	\\
	\text{output gate parameters:}	\quad\quad &
		\begin{bmatrix} \mathbf{W_{xo}} \\ \mathbf{W_{ho}} \end{bmatrix}  &
		\in \mathbb{R}^{(D+H)\times H},\, \mathbf{b_o}\in \mathbb{R}^H	\\
	\text{network output layer parameters:}	\quad\quad &
		\mathbf{W_{hK}} & 
		\in \mathbb{R}^{H\times K},\, \mathbf{b_K}\in \mathbb{R}^K
\end{eqnarray}
where $D$ is the dimension of the input signal, $H$ is the number of LSTM units, and $K$ denotes the number of classes (in our case: activities).
$\mathbf{W}_{(\cdot)}$ are weight matrices and $\mathbf{b}_{(\cdot)}$ are bias vectors that define the transformation of the particular gate as illustrated in Fig. \ref{fig:LSTM_unit}. 
Given the input and model parameters, the forward pass can be written as follows:
\begin{eqnarray} \label{eq:lstmFF}
 \mathbf{f}_t & = & \sigma \left(\begin{bmatrix} \mathbf{W_{xf}} \\ \mathbf{W_{hf}} \end{bmatrix}^T [\mathbf{x}_t, \mathbf{h}_{t-1}] +  \mathbf{b_f} \right)\\
 \mathbf{i}_t & = & \sigma \left(\begin{bmatrix} \mathbf{W_{xi}} \\ \mathbf{W_{hi}} \end{bmatrix}^T [\mathbf{x}_t, \mathbf{h}_{t-1}] +  \mathbf{b_i} \right)\\
 \tilde{ \mathbf{c}_t} & = & \tanh \left(\begin{bmatrix} \mathbf{W_{xc}} \\ \mathbf{W_{hc}} \end{bmatrix}^T [\mathbf{x}_t, \mathbf{h}_{t-1}] +  \mathbf{b_c} \right)\\
 \mathbf{c}_t & = & \mathbf{f}_t\circ \mathbf{c}_{t-1}+\mathbf{i}_t \circ  \tilde{ \mathbf{c}_t}\\
 \mathbf{o}_t & = & \sigma \left(\begin{bmatrix} \mathbf{W_{xo}} \\ \mathbf{W_{ho}} \end{bmatrix}^T [\mathbf{x}_t, \mathbf{h}_{t-1}] +  \mathbf{b_o} \right)\\
 \mathbf{h}_t & = &\mathbf{o}_t \circ \tanh (\mathbf{c}_t)
\end{eqnarray} 
where $\mathbf{f}_t, \mathbf{i}_t,  \tilde{ \mathbf{c}_t}$, and $\mathbf{o}_t$ denote outputs of the forget, input, cell, output gates, respectively (inside LSTM units), {\color{black}and $\circ$ stands for element-wise multiplication}.
$\mathbf{c}_t$ and $\mathbf{h}_t$ are the outputs of the LSTM units, 
and can be passed to the next timestep to iterate the aforementioned process.
Given $\mathbf{h}_t$,  
prediction can be performed 
and the class probability vector $\mathbf{p}_t = [p_{t1}, p_{t2}, ..., p_{tK}] \in \mathbb{R}^K$ is calculated as follows:
\begin{equation}\label{eq:softmax}
\mathbf{p}_t = s(\mathbf{W_{hK}}^T\mathbf{h}_t + \mathbf{b_K}),
\end{equation}
where $s(\cdot)$ is a softmax function.
The class label $\hat{k_t}$ at time $t$ is then assigned to the one with the highest probability, i.e., 
\begin{equation}\label{eq:assign_label}
\hat{k_t}=\underset{[1,K]}{\arg\max}\, p_{tk}.
\end{equation}
%
Throughout this paper we refer to an LSTM model (incl.\ model parameters) as $\mathbf{W}_{LSTM}$.

\begin{figure}[tp]
	\subfloat[[CNN for frame-based HAR (c/o \protect\cite{Zeng:2014tk}).]{
		\includegraphics[height=3.5cm]{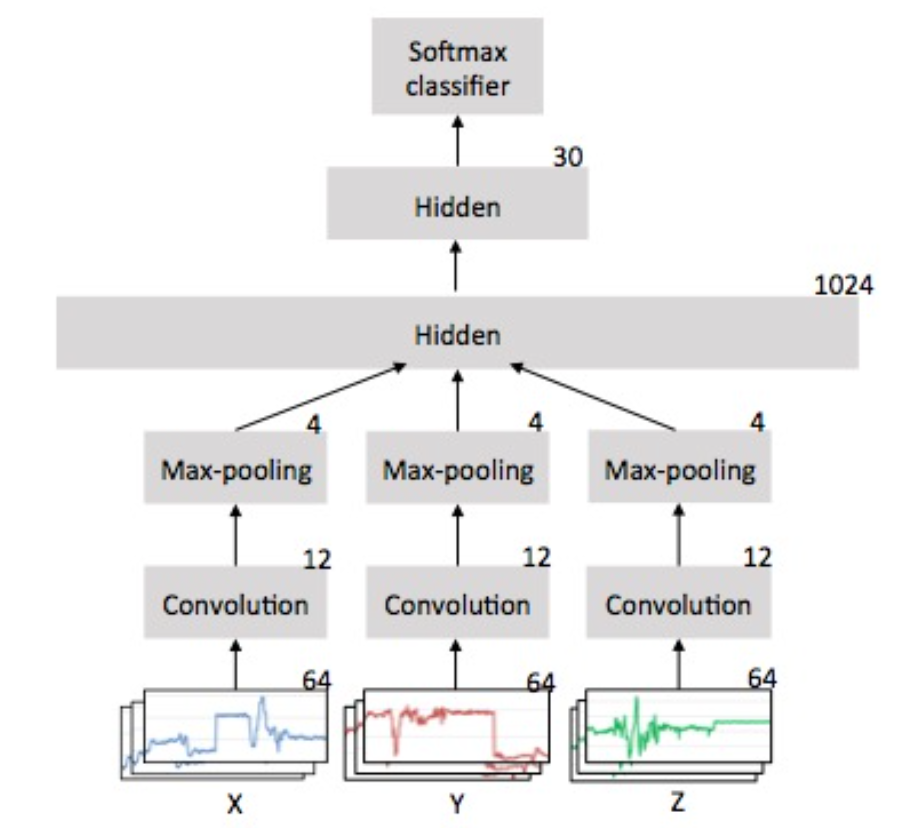}
		\label{fig:deepLearning:CNN}	
	}
	\hfill	
	\subfloat[Deep LSTM networks for HAR (c/o \protect\cite{Hammerla2016}).]{
		\centering
		\includegraphics[height=3.5cm]{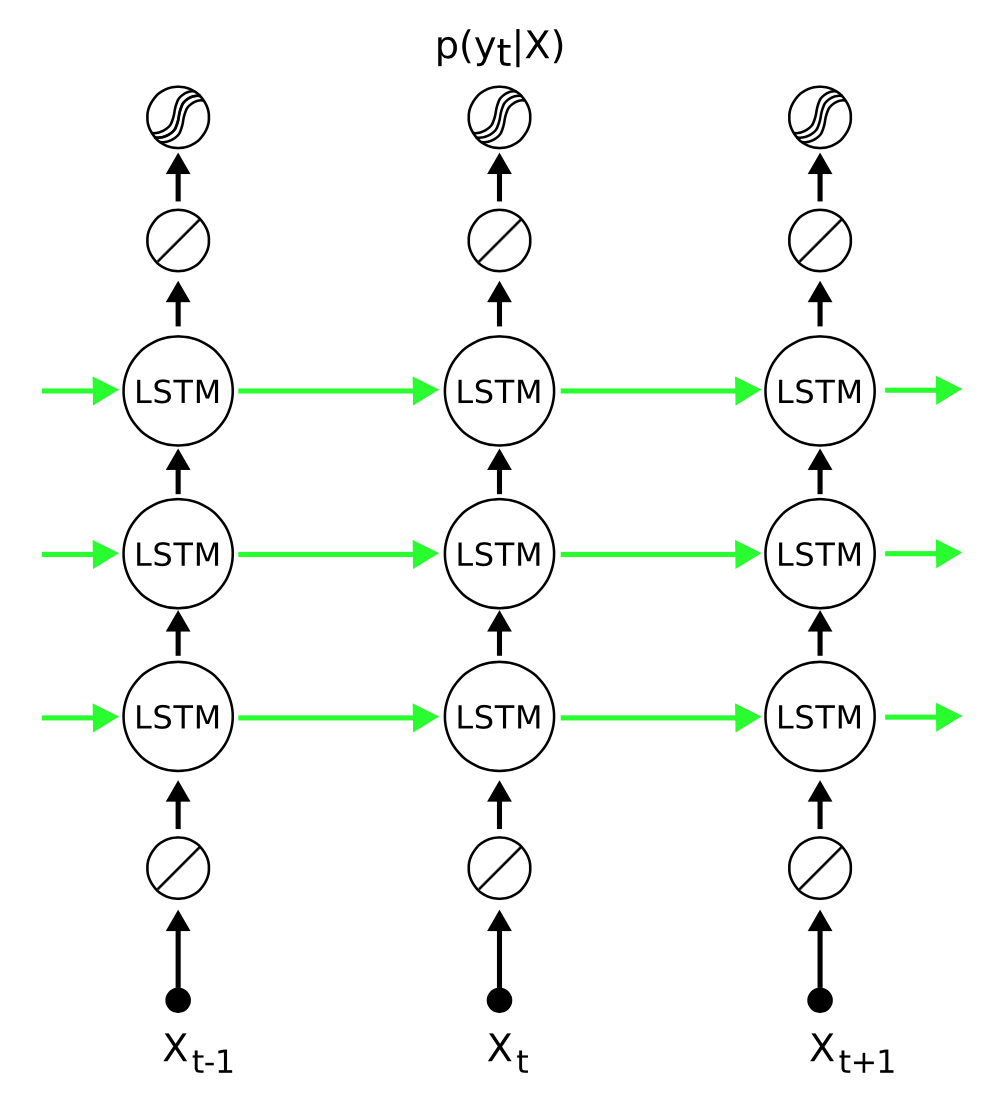}
		\label{fig:deepLearning:LSTM}	
	}
	\hfill	
	\subfloat[Combinations of CNN and LSTM (c/o \protect\cite{morales2016deep}).]{
		\includegraphics[height=3.5cm]{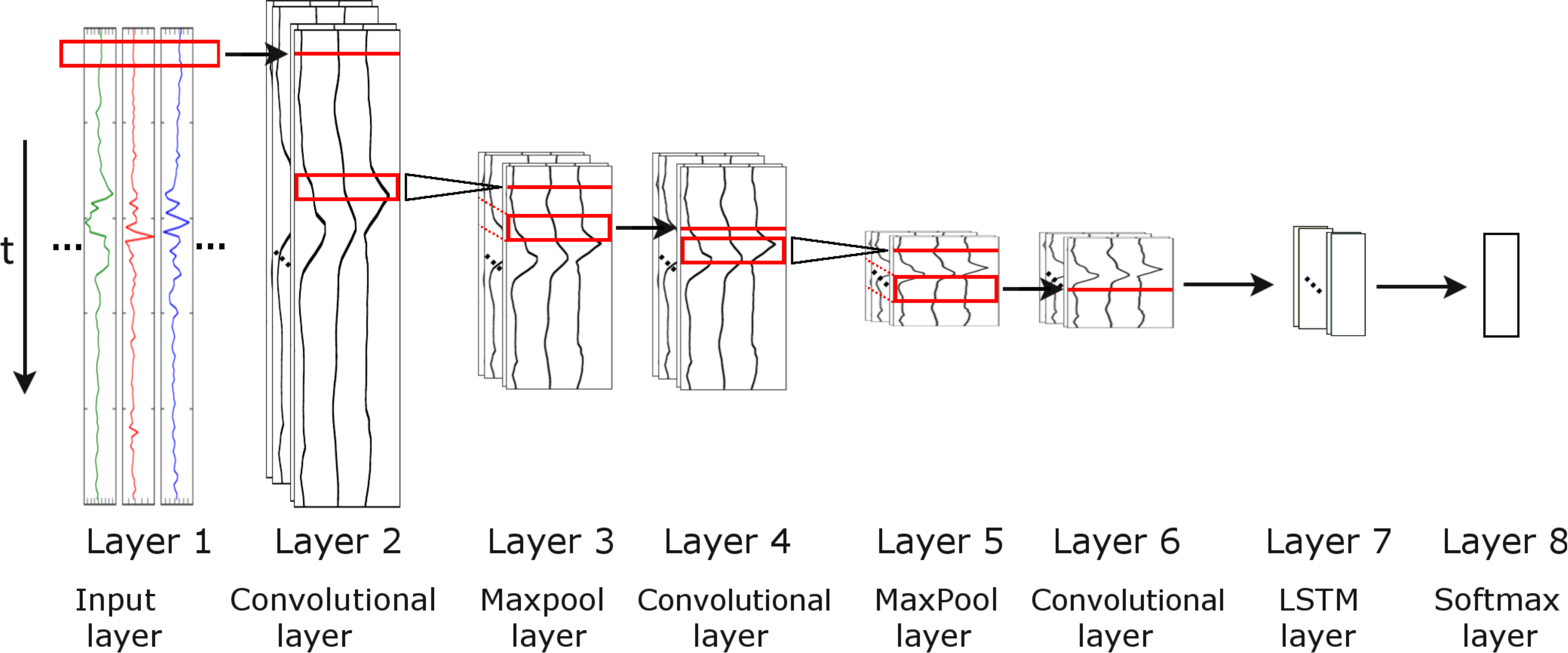}
		\label{fig:deepLearning:CNNnLSTM}	
	}
	\caption{Deep learning for Human Activity Recognition as it has been used in ubiquitous and wearable computing (selection).}
	\label{fig:deepLearningHAR}
\end{figure}

Fig.\ \ref{fig:deepLearningHAR} gives an illustrative overview of the most prominent deep learning approaches for human activity recognition using wearable sensing platforms.
It shows examples of:
\textit{i)} CNN-based HAR systems (Fig.\ \ref{fig:deepLearning:CNN});
\textit{ii)} LSTM-based approaches to HAR (Fig.\ \ref{fig:deepLearning:LSTM}); and
\textit{iii)} combinations of the two (Fig.\ \ref{fig:deepLearning:CNNnLSTM}).
Especially the latter is of particular interest as it combines the effectiveness of CNN-based representation learning with sequence modelling capabilities of LSTMs.
Such a combination has been used very successfully for activity recognition \cite{Ordonez2016}, but also for the exploration of transfer learning scenarios using deep learning \cite{morales2016deep}.

\subsection{Ensembles of Classifiers}
\label{sec:related:ensembles}
The bulk of general machine learning research is directed towards the development of effective classification methodologies that can be applied to various recognition tasks.
Thereby the focus is typically on pushing the boundaries for individual classifiers to be employed for modelling and inference.
In addition to this, in recent years the combination of multiple learners into meta-classifiers has gained popularity as it has been shown that such classifier Ensembles can be very beneficial for the analysis of complex datasets \cite{Kittler1998-OCC}.
Typically, a range of different base classifiers is estimated either on modified training sets or on alternative representations of the training data. 
During inference the whole set of classifiers is applied to the original recognition task, which results in multiple predictions that are then aggregated in order to achieve a final classification. 
Multiple variants have been developed for both generating different views on the training data, and for aggregating individual predictions.
It has been shown that classifier Ensembles have great potential to outperform traditional single classifier systems especially on challenging recognition tasks \cite{kuncheva2004combining}.

In the original formulation of the concept of classifier Ensembles the principle constraint for base classifiers was rather modest, requiring classification accuracy of Ensemble members to just be better than random -- so-called weak learners form the classifier Ensembles.
However, in recent years it has been shown that Ensemble classification is even more effective when strong base classifiers are integrated (e.g., \cite{Kim:2003co}).
Either way, Ensemble approaches will only lead to improved performance if the underlying base classifiers capture substantial, mutual diversity, that is, if modelling different characteristics of a data set.

A prominent approach to Ensemble creation utilises artificially constrained training sets for base learner training.
Through repeated random selection of samples from the training data the so-called Bagging procedure aggregates classifiers that are estimated on bootstrap replicates of all training samples \cite{Breiman1996-BP}. 
The popular Boosting procedure pursues an alternative approach to generating diversity in classifier Ensembles, namely through targeted (re-)weighting of sample data for their consideration into the training procedure. 
In an iterative approach Boosting gradually focuses more on those samples that are harder to classify, i.e., causing higher classification errors \cite{Schapire:1990cg}.

Arguably, deep learning methods can already be interpreted as Ensemble learning methods.
Through the combination of multiple hidden layers, each containing numerous processing nodes that eventually contribute to an aggregated classification decision (at the output layer of a network) these models effectively resemble the principle of Ensemble learning.
However, the focus of this paper is at a higher level of abstraction, that is, on combining multiple \textit{complete} networks to tackle challenging recognition tasks.
So far there has only been little work on Ensembles of deep learning methods.
To the best of our knowledge this paper is the first that addresses HAR using Ensembles of  recurrent LSTM networks.

Examples of deep learning Ensembles include the combination of Support Vector Machines (SVM) into classifier Ensembles \cite{Qi:2016ke}.
It has been shown that the resulting stacked "Deep SVMs" outperform non-Ensemble approaches in a variety of standard classification benchmarks -- not including time series analysis such as activity recognition though.
Furthermore, stacking methods for Ensemble learning of sets of deep neural networks have been applied successfully to speech recognition applications where class posterior probabilities as computed by convolutional, recurrent, and fully-connected deep neural networks are integrated into an Ensemble \cite{Deng:2014vj}.

The focus of this work is on combining deep learning methods, that are --to date-- most effective for human activity recognition, into classifier Ensembles.
Specifically we focus on LSTM networks as base learners given their superior performance in challenging HAR benchmarks \cite{Ordonez2016,Hammerla2016}.
In line with previous work on combining strong learners into Ensembles, we hypothesise that the combination of diverse individual LSTM networks will lead to more robust and thus improved activity recognition results in challenging Ubicomp scenarios.

{\color{black}Our hypothesis is supported by (few) recent related works from other application domains where LSTM models have been integrated into classifier collectives leading to improved recognition results, which is encouraging for real-world HAR.
Most related to our work, \cite{Song2016-DEL} have described a system that utilises Ensembles of LSTM learners for predicting actions from video data. Unlike our approach their Ensemble is based on a range of base learners that are derived specifically for variants of certain activities, that is the diversity required for building Ensembles is de-facto prescribed. We argue that for real-world HAR scenarios such a prescription represents an unnecessary and potentially harmful limitation with regards to flexibility and robustness of the resulting classifier collectivce. As such, whilst our general argumentation for using strong --LSTM-- base learners in Ensembles is identical, we focus on data-driven diversity that leads to effective and robust recognition systems.
Other authors have referred to LSTM Ensembles as well, yet with a different --somewhat misguiding-- interpretation. For example, \cite{Singh2015-EOD} employed three variants of LSTM networks (vanilla deep LSTM, and two variants of bi-directional LSTMs) in parallel for biological sequence analysis and then applied majority voting for final classification. Whilst this approach touches upon the general idea of classifier Ensembles \cite{Kittler1998-OCC}, it does so in a very constrained way, which leaves substantial room for improvement. The focus of our work is on data-driven generation of diversity through a novel training procedure that allows for deriving very powerful and especially robust LSTM ensembles as they are of relevance for real-world HAR scenarios using wearable and pervasive sensing modalities.
}

%% file: system.tex

\section{Ensembles of Deep LSTM Learners for Activity Recognition}
\label{sec:system}
Our work is mainly motivated by the following observations.
Activity recognition based on mobile and wearable sensing modalities --as it is typical for the field-- often face substantial challenges with regard to data quality and, to some extent, quantity.
Firstly, raw sensor data are inherently noisy.
High temporal resolutions of sensing facilities used for \textit{direct} movement sensing in principle results in substantial variations of raw sensor data even within the same set of activities, i.e., ambiguous representations.
Additionally, faulty sensor operation not only occasionally leads to erroneous sensor readings that further challenge the quality of the data.
Secondly, the widespread adoption of mobile computing and their integrated sensing facilities now render the collection of raw, unlabeled data an almost trivial endeavour.
However, the collection of ground truth annotation is substantially more challenging -- primarily for practical as well as for ethical (privacy) reasons. 
It is this discrepancy that typically leads to rather imbalanced datasets with non-uniform class distributions, which represents considerable problems for many training and validation approaches.

Thirdly, even though the popular sliding window, frame-based analysis approach effectively circumvents the segmentation challenge, that is the need for locating (within a stream of sensor data) episodes of relevant activities before classifying them, fine grained analysis of complex activities substantially suffers from such a simplification.
Strictly speaking, sequential data shall be modelled by and analysed with sequential models.
With growing maturity of the field of HAR and increasingly more complex recognition tasks the research community is more and more acknowledging this problem.
As demonstrated in recent related work (e.g., \cite{Hammerla2016,morales2016deep}) deep, recurrent neural networks, i.e., sequential models, have great potential for robust analysis beyond the predominant sliding window, frame based analysis paradigm.


In response to these observations and with the aim of improving the robustness of activity recognition systems we have developed a HAR framework based on Ensembles of  LSTM learners.
We combine multiple deep, recurrent neural networks into a meta classifier that is substantially more robust w.r.t.\ the aforementioned challenges in ubiquitous and wearable computing applications of human activity recognition.
Fig.\ \ref{fig:system} gives an overview of our system, which will be described in detail in the subsequent sections.
Our approach is characterised by the following key aspects:

\begin{figure}[tp]
	\vspace*{-1em}
	\centering
	\includegraphics[width=0.95\textwidth]{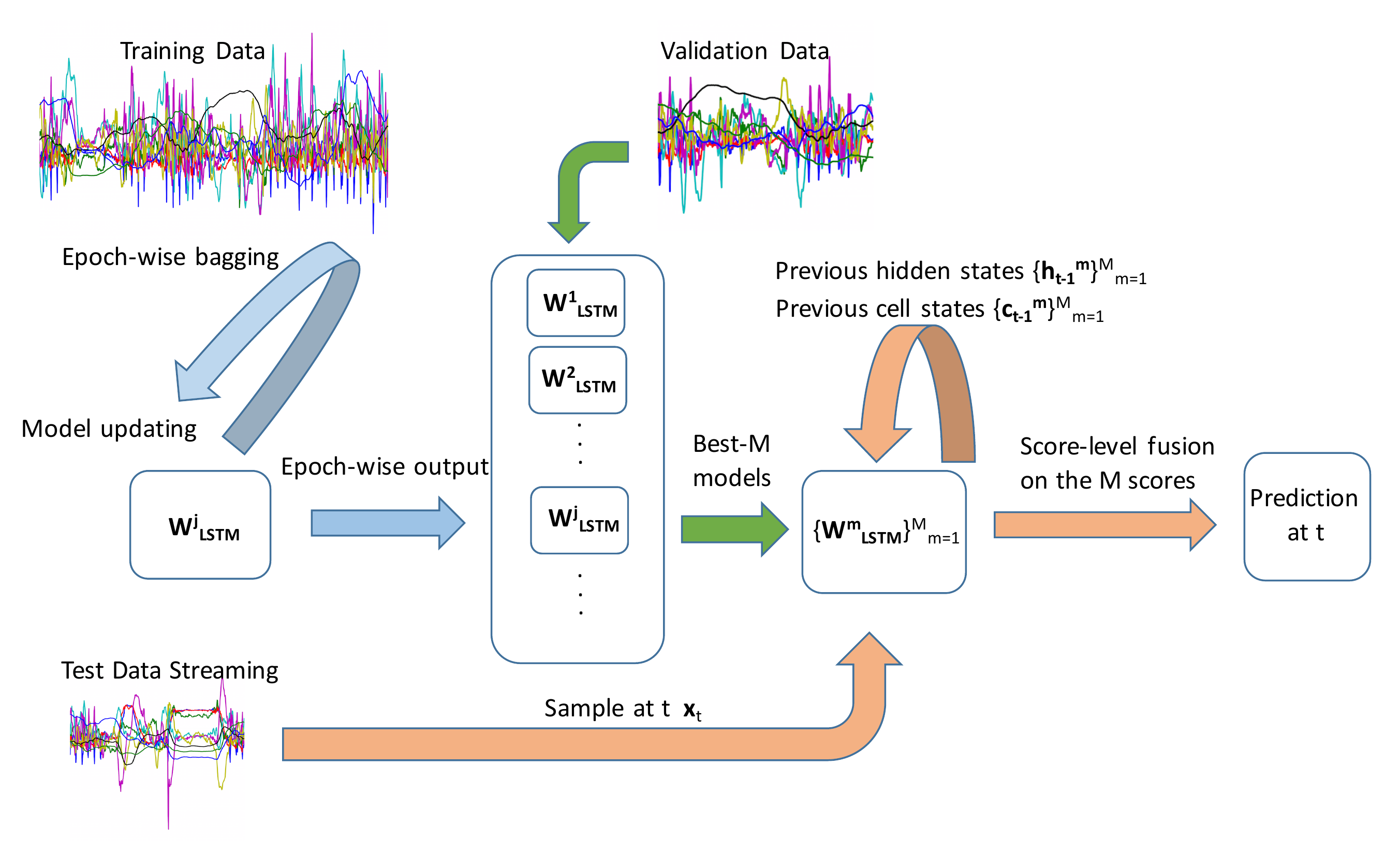}
	\vspace*{-1em}
	\caption{Ensembles of deep LSTM learners for activity recognition -- system overview.}
	\label{fig:system}	
\end{figure}

\paragraph*{LSTM based HAR}
Deep learning has successfully been used in a variety of application domains, including HAR using ubiquitous and wearable sensing. 
In line with recent related work (e.g., \cite{Ordonez2016,Hammerla2016}) we employ a specific variant, namely deep, recurrent networks based on Long Short Term Memory (LSTM) units \cite{Hochreiter1997}.

\paragraph*{Modified Training and Sample-Wise Model Evaluation}
The majority of related work in HAR utilises (variants of) the well established sliding-window, frame-based analysis approach \cite{Bulling2014}.
To some extent this also holds for the emerging field of deep learning based HAR where portions of contiguous sensor readings --frames that are shifted along the time series data with substantial overlap between subsequent frames-- are analysed by the complex networks and predictions are given at the level of frames rather than samples.
We use an alternative approach, namely random sampling of frames with varying length during model training, and sample-wise model evaluation during inference.

\paragraph*{Fusion of multiple LSTM learners into Ensembles}
Based on the modified training procedure we are able to train models that are more robust with regards to aforementioned data challenges.
We generalise this concept through repeated iteration and combine multiple models that have been derived in this manner into Ensembles of LSTM activity recognition models, which effectively resembles the general concept of Bagging \cite{Breiman1996-BP} but now with substantially more complex base learners.

{\color{black} 
\subsection{Motivation for Fusing LSTM Learners}
\label{sec:system:motivation}
With regards to error types, an ideal classifier should have low bias, and low variance. 
Bias here corresponds to the expected error between predicted values and the ground truth, whereas variance measures the variability of model prediction for a given data point.
Real world recognition systems need to find a compromise / define a tradeoff between bias and variance, and the classifiers may suffer from under- (high bias) or overfitting (high variance) -- depending on representativeness of training data and classifier complexity \cite{Hastie:2009wp}.
For example, simple linear classifiers (such as logistic regression) may have high bias (i.e., underfitting) when facing non-linear data.
Complex non-linear models, such as neural networks, empirically tend to have low bias (i.e., low training errors), yet they tend to overfit when training datasets are not large / representative enough (higher variance).  
In contrast to bias reduction techniques (such as boosting) for simple models \cite{baggingBoosting2012}, Bagging (i.e., Bootstrap Aggregating) \cite{Breiman1996-BP} is a variance reduction technique for strong classifiers.
Based on bootstrapped training samples, strong base classifiers are initially trained with high variance, which can later be reduced through aggregation, e.g., model averaging.
The most popular strong base learners include decision trees \cite{RF2001}, neural networks \cite{NNEnsemble2002}, or kernel methods \cite{kernelEnsemble2004}, and the corresponding variances are significantly reduced through bagging, yielding improved generalisation capabilities.

Motivated by these generic considerations our recognition system is based on Ensembles of LSTMs.
Within the general field of pattern recognition and machine learning it has been shown that individual classifiers can be integrated into Ensembles and used very successfully for solving challenging recognition tasks (e.g., \cite{Kittler1998-OCC,kuncheva2004combining}).
Despite the fact that the general machine learning literature largely argues w.r.t.\ static classifiers, there is no  reason for this observation not to be valid for sequential models as well (cf.\ Appendix \ref{app:loss} for a more formal exploration).
}

{\color{black}
Consequently, we employ strong learners --LSTMs, which have low bias and high variance owing to their almost universal function approximation capabilities-- for deriving our recognition Ensembles. 
We follow the previously described general pipeline, that is we bootstrap strong learners, followed by variance reduction through model averaging.
Since computational complexity can quickly become an issue when training deep neural networks, such as LSTMs, we employ an epoch-wise bagging scheme. 
That is, our scheme generates one LSTM base learner per epoch.
We have further modified the training procedure by randomising various parts of it (as described below) aiming at increased diversity amongst base learners, which is crucial for Ensembles. 
Through aggregating the resulting base learners, we effectively counteract the high variance problem, i.e., the notorious tendency of neural networks to overfit. 
Most importantly, the developed approach is effective for sequential base learners (LSTMs in our case), which is advantageous for the analysis of time-series data as it is relevant for HAR settings.
}


%
%
%

\subsection{Modified Training for LSTM Models}
\label{sec:system:LSTM}
\begin{figure}[tp]
	\centering       	
      	\includegraphics[width=16cm, height=6cm]{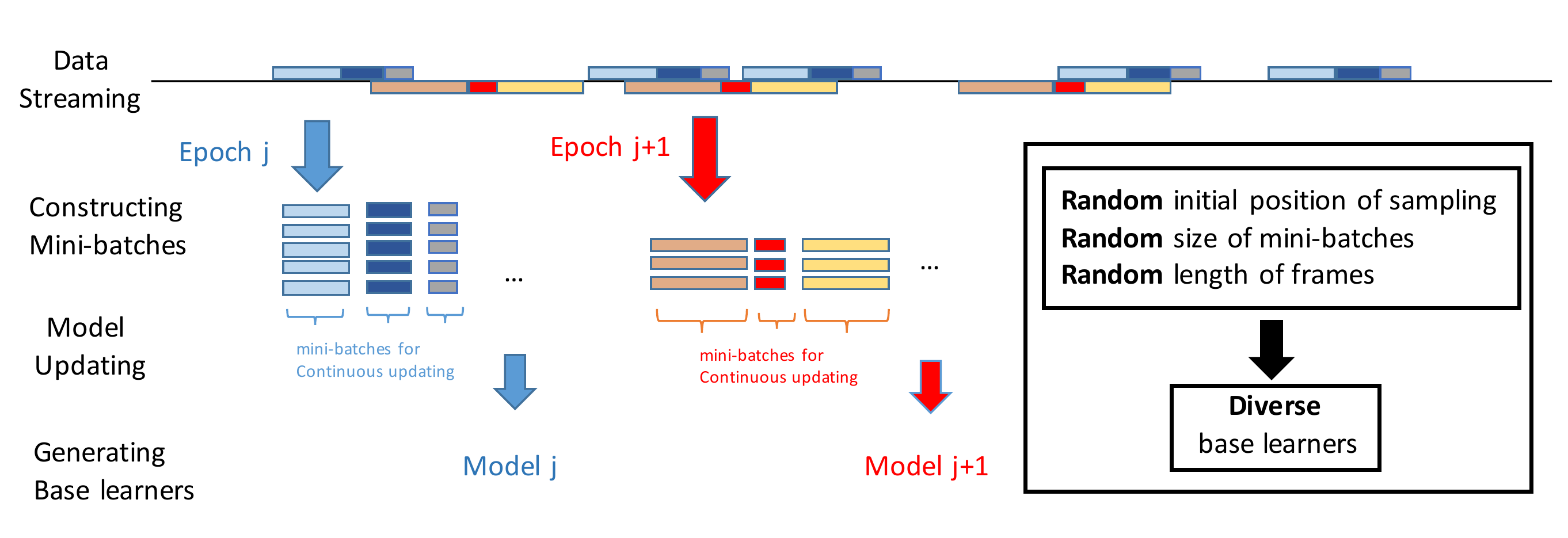}
      	\caption{{\color{black} Illustration of the epoch-wise bagging developed in this paper (formal description in Alg.\ \ref{alg:epochBagging}). }

} 
	\label{fig:bagging}	
\end{figure}

Our activity recognition framework is based on Long Short Term Memory (LSTM) networks as summarised in Sec.\ \ref{sec:related:LSTM} \cite{Hochreiter1997}.
Whilst we employ standard LSTM models at inference time, our training procedure is different as it is targeted to meet challenging real-life data scenarios as they are typical HAR for the field of ubiquitous and wearable computing.
Fig.\ \ref{fig:bagging} illustrates the developed training procedure for individual LSTM learners and Alg.\ \ref{alg:epochBagging} summarises the procedure formally (cf.\ Sec.\ \ref{sec:related:LSTM} for generic nomenclature).

\subsubsection{Epoch-wise Bagging}
\label{sec:system:LSTM:epochBagging}
As per the aforementioned discussion our hypothesis is that model training procedures in HAR are challenged by both noise in the sensor readings as well as class imbalance.
Unfortunately, in real-world scenarios it is typically not straightforward to formalise these challenges w.r.t., for example, quantifying the percentage of problematic sensor readings let alone identifying them \textit{before} having access to recognition results.
In our work we acknowledge these challenges and do not attempt to solve this "chicken-and-egg" problem nor do we require prior knowledge about the challenges of a specific application domain.
Instead, we assume that a certain percentage of the sample data is at least not beneficial if used for model training and thus it will not be used.
Without prior domain knowledge or any other constraints we do not know which portions of the data are of inferior quality for the model training.
Aiming for generalisation we thus employ a method that creates the effective training set through probabilistic selection of subsets of the original data and use these for mini-batch based learning of LSTM networks using stochastic gradient descent (SGD). 
The method presented here extends our previous work \cite{Hammerla2016}.

Stochastic gradient descent is a standard method for learning generalised deep models \cite{Bottou:2010br}.
For i.i.d.\ samples, such as images, the standard procedure is to shuffle the whole training set before each epoch and then to generate subsets of the shuffled training data --so-called mini-batches-- to be used for model training.
The gradient (corresponding to each mini-batch) can then be calculated by minimising an error (also referred to as loss) function for model updating. 
However, as it has been shown before due to the contextual nature of time series data, this uninformed shuffling cannot directly be applied \cite{Hammerla:2015vj}.
In order to construct diverse yet usable mini-batches for training models for time series analysis, in our previous work we have randomly chosen $B$ different start positions $\{q_0^b\}_{b=1}^B$ for generating $B$ frames of contiguous portions of sensor data along the whole training sequence \cite{Hammerla2016}.
This procedure is repeated for every epoch of model training.
Mini-batches of every training epoch are constructed by selecting frames of sensor data with pre-defined window length $L$ from the chosen $B$ starting positions within the stream of sensor data.
Formally, $B$ frames of contiguous sensor readings are selected with start/end indices defined as follows:
\begin{equation}
	 	 \bigg \{ \quad   [q_i^b, q_{i+1}^b] \quad   \bigg | \quad    q_{i+1}^b-q_i^b = L, \quad i=0, 1, 2, ..., i^{max}, \quad   q^b_{i^{max}}-q^b_0 \approx T       \bigg \}_{b=1}^B,
	 \label{eq:minibatchIJCAI}
\end{equation}
where $T$ denotes the number of samples within the training data.
For training data $\mathbf{X}\in \mathbb{R}^{D\times T}$, the $i$-th mini-batch $\mathbf{g}_i$ is then created as follows:
\begin{eqnarray}
	\mathbf{g}_i & = & \{\mathbf{X'}[:, q_i^b: q_{i+1}^b] \}_{b=1}^B\\
	\text{where} \quad \mathbf{X'} & = & [\mathbf{X}, \mathbf{X}]\in \mathbb{R}^{D\times2T}.
\end{eqnarray}
This procedure was originally referred to as "wrapping around" procedure \cite{Hammerla2016}.


For increased robustness of the resulting HAR models w.r.t.\ aforementioned challenges of real-world scenarios, 
in this work we extend 
the wrapping around procedure into a Bagging scheme for epoch-wise learners.
Instead of using pre-defined, fixed numbers of training frames $B$ (i.e., mini-batch size) and frame length $L$  --for which it is difficult to find a justification when no or only little prior knowledge about a specific domain is available at modelling time--  we randomly choose $\hat{B}$ 
for each epoch, and within each  epoch we also randomly sample various frame lengths $\hat{L}$. 

{\color{black} 
It is worth mentioning that our general LSTM training scheme is insensitive to specific window lengths (within a range) because the general capability of LSTMs to "remember" contextual temporal information remains unchanged.
LSTMs pass their current (internal) cell state and hidden state as inputs to the next time-step.
We avoid hard decisions regarding the length of the contextual window that is being used during a training step because it is difficult to make such a generalised decision upfront. 
Practically all of related work is based on fixed window lengths, which essentially represents some sort of compromise based on heuristics. 
Instead we set up various, randomly chosen window lengths as well as randomised starting points / mini-batch sizes with the specific goal to increase the diversity for epoch-wise classifiers.
Through repeating the randomised process (over a number of epochs), we make sure that --empirically-- we cover the whole sequence of input data.
} 

{\color{black}
However, for each epoch}
our mini-batches only cover contiguous signal parts of length $\lfloor T/\hat{B} \rfloor$.
Overall, only a total of $\hat{B} \lfloor T/\hat{B} \rfloor$ samples are used in each epoch. 
Specifically, 
$\hat{B}$ frames of contiguous sensor readings are generated with start/end indices defined as follows:
\begin{equation}
\bigg \{ \quad   [q_i^b, q_{i+1}^b] \quad   \bigg | \quad    q_{i+1}^b-q_i^b = \hat{L}_i, \quad i=0, 1, 2, ..., i^{max}, \quad   q^b_{i^{max}}-q^b_0 = \lfloor T/\hat{B} \rfloor   \bigg \}_{b=1}^{\hat{B}},
	 \label{eq:minibatch_ours}
\end{equation}
where $\hat{B}$ and $\hat{L}_i$ are generated using a discrete, uniformly distributed random number generator.  

Inevitably, this procedure results in repetitions and/ or omission of parts of the original training data -- which, according to our aforementioned motivation, is key for robust model training in real-world scenarios.
In our practical experiments we found that on average about  $35\%-38\%$ of the training data is not used per epoch -- when using a standard, i.e., uniformly distributed random number generator for defining the training frames.
Given that these omitted portions are selected probabilistically this method has the --intended-- positive effect of yielding less correlated epoch-wise classifiers.
Essentially, this concept resembles the {\color{black} aforementioned idea}
of Bagging \cite{Breiman1996-BP} -- but at the level of {\color{black} bootstrapping for epoch-wise LSTM learners} and without specific model score combination.

\begin{algorithm}[tp]
\SetKwInput{KwData}{Input}
\SetKwInput{KwResult}{Output}
\KwData{training data: $\mathbf{X}\in \mathbb{R}^{D\times T}$} 
\KwResult{ $\{\mathbf{W}_{LSTM}^j\}_{j=1}^{MaxEpoch}$  -- the whole set of individual LSTM learners} 

\vspace*{1em}
initialise the model $\mathbf{W}_{LSTM}^0$ ;

\vspace*{1em}
\For{$j  = 1$ \KwTo $MaxEpoch$}{
draw random integer $\hat{B}_j$ (i.e., mini-batch size) from discrete uniform distribution $\textit{U}(B_{low}, B_{high})$\;

draw $\hat{B}_j$ random integers $\{q_0^b\}_{b=1}^{\hat{B}_j}$ (i.e., starting positions) from discrete uniform distribution $\textit{U}(1, \lfloor T(1 - \frac{1}{\hat{B}_j})\rfloor)$\;

initialise:  $i=0, \quad L_{seq}=0$ \; 

\While{$L_{seq} \le \lfloor T/\hat{B}_j \rfloor$}{
	draw random integer $\hat{L}_i$ (i.e., frame length) from discrete uniform distribution $\textit{U}(L_{low}, L_{high})$\;
         $ \{q_{i+1}^b = q_i^b+\hat{L}_i \}_{b=1}^{\hat{B}_j}$\;
	construct the current mini-batch $ \mathbf{g}_i^j $ such that
		 $ \mathbf{g}_i^j = \{\mathbf{X}[:, {q_i^b:q_{i+1}^b}]\}_{b=1}^{\hat{B}_j}$  (also see Fig. \ref{fig:bagging}) \;

	 update the model $\mathbf{W}_{LSTM}^{ji}$ using the calculated gradient $\delta(\mathbf{g}_i^j $) based on a certain loss function (e.g., cross entropy loss)\;
	 
	 $i = i + 1$  \;	
	  $L_{seq} = L_{seq}  + \hat{L}_i$ \; 
 }
 output $\mathbf{W}_{LSTM}^{j}$ for the $j^{th}$ epoch
}
 \caption{Epoch-wise bagging for LSTM learners. }
 \label{alg:epochBagging}
\end{algorithm}

{\color{black} 

\subsubsection{Alternative Loss Function for LSTM learners}
\label{sec:system:LSTM:loss}

In addition to employing the most popular cross entropy loss (CE) function (more details can be found in Appendix \ref{app:loss}), we also explore the utility of an alternative loss function. 
Specifically, we train base learners using the $F1$-score loss, which is a cost-sensitive loss function that is negatively correlated to the approximation of the $F1$ score for a dataset (cf.\ \cite{Dembczynski:2013wx} for a general introduction).
Different from CE, it is a global cost function for the complete dataset and it cannot directly be used to measure individual sample's losses \cite{Fmeasure}.
For a set of $N$ samples,  $F1$ loss is defined as:      
\begin{equation}
	\label{eq:avgLossF1}
	L^{F1}=   \mathbb{E}_{k\in [1,K]}\bigg \{ 1- \frac{2\sum_{t=1}^N\mathbf{p}_t \circ \mathbf{z}_t}{\sum_{t=1}^N\mathbf{p}_t +\sum_{t=1}^N\mathbf{z}_t }\bigg \},
\end{equation}
where $\circ$ denotes element-wise multiplication, $\mathbf{p}_t$ is the probabilty vector at time step $t$ (Eq.\ \ref{eq:softmax}), and $\mathbf{z}_t \in \mathbb{R}^K$ defines the binary vector indicating correct class label $k$ such that $z_{tk}=1 \land \{z_{tj}=0\}_{j\neq k}$.

Although $F1$ loss considers the data imbalance problem, LSTM learners are still trained based on mini-batches (for efficiency and numerical stability reasons), which may not precisely approximate the "true" gradient towards an optimal $F1$ score over the whole training set -- but a reasonable approximation.
Owing to its aforementioned properties, the classifiers generated by F1 loss may have different behaviours compared to their CE counterparts, and we expect it may benefit the overall performance by fusing LSTM classifiers from these two sources.


} 

\subsection{Combining Multiple LSTMs into HAR Ensembles}
\label{sec:system:ensemble}
The modified training procedure for individual LSTM learners leads to 
various models due to varying the "view" on the training data.
As mentioned before this effectively resembles the main concepts of Bagging, i.e., a --simplified-- variant of Ensemble learning.
The general idea of Ensemble classifiers is to create collectives of individual learners that are each trained on different views of the sample data.
Through introducing variation, individual learners focus on different aspects of a problem domain, which in combination leads to more robust and typically better recognisers in terms of average classification accuracy.

With the concept of epoch-wise bagging as described in the previous section, we already have introduced a concept for diversifying the training of our activity recognition system.
We now generalise this idea by combining multiple LSTM models into classifier committees.
For doing so we focus on generating diversity from the two main sources discussed in the previous sections:
\begin{enumerate}
\item Diversity of base learners due to random selection of subsets of training data for parameter estimation. 
Essentially this corresponds to combining the individual LSTMs obtained from every training epoch. 

\item Diversity of LSTM models caused by employing two different loss functions for model training,
which leads to variations in resulting LSTM models.
\end{enumerate}

Independent of the cause for diversity in base classifiers (in our case LSTMs) the meta-optimisation objective for classifier Ensembles is to generate classifiers that are diverse and minimally correlated \cite{Kittler1998-OCC}.
Actual model combination is based on score level fusion as explained below.

According to Alg.\ \ref{alg:epochBagging} our training procedure results in $MaxEpoch$ LSTM models $\{\mathbf{W}_{LSTM}^j\}_{j=1}^{MaxEpoch}$.
Typical for deep learning approaches, $MaxEpoch$ is thereby often a large number ($MaxEpoch=100$ in this work). 
Any two models $\mathbf{W}_{LSTM}^j$ and $\mathbf{W}_{LSTM}^{j+1}$ may be marginally correlated only due to the substantial variability in the training sets generated for the particular epochs $j$ and $j+1$. 
Our training procedure now analyses the quality of all $MaxEpoch$ models for the separate validation set.
Based on these results we then only preserve the $M$ best models for inclusion into the overall Ensemble -- through score level fusion.
It is worth reiterating that we employ our Ensembles of LSTM learners for sample-wise activity recognition.
For every sample, i.e., every time step $t$, the final softmax layer of each of the $M$ models provide class probability vectors $\mathbf{p}_t^m$ as defined in Eq. (\ref{eq:softmax}).
Class probability vectors effectively describe probability distributions over the classes (activities) of interest for a specific task.
Probability vectors from all $M$ models are then combined via their arithmetic means resulting in a final score vector $\mathbf{p}_t^{fusion}$:

\begin{equation} 
	\label{eq:fusion}
	\mathbf{p}_t^{fusion} = \frac{1}{M}\sum_m^M{\mathbf{p}^{m}_t}.
\end{equation} 

Fusing the $M$ best performing and typically least correlated LSTM learners leads to substantially improved model performance.
{\color{black}Appendix \ref{app:loss} provides a more formal exploration of the general concept of integrating LSTM base learners into Ensemble classifiers.}

%% file: experiments.tex
\section{Experimental Evaluation}
\label{sec:experiments}
In order to validate the effectiveness of our HAR framework 
we conducted an experimental evaluation.
For this we ran recognition experiments on three benchmark datasets that are considered standard in the field 
and thus widely used in the literature. 
In what follows we will first describe our training and evaluation methodology, followed by a summary of the details of the datasets used for the evaluation study, and finally present and discuss our results.

\subsection{Model Training, and Evaluation Protocol}
\label{sec:experiments:protocol}
All experiments were conducted using the framework introduced in this paper.
As such they share a set of common parameters that are described in what follows thereby covering general model configurations, as well as training and evaluation methodologies as explained below.
Dataset specific modifications --if any-- are discussed in the dataset description subsections.
Apart from potential subsampling and modality fusion, sensor data are used "as is" that is without employing feature extraction methods be it based on heuristics \cite{Ploetz:2010p8026,Bulling2014} or on more sophisticated transformations \cite{Hammerla:2013wo}, which is in line with the majority of recent, Deep Learning based analysis methods in the field of human activity recognition using wearable sensing platforms.

\subsubsection{Model Configuration}
\label{sec:experiments:models}
For all experiments that use our LSTM learners, i.e., those based on individual learners as baselines as well as those employing the Ensembles, we use the following LSTM model configuration.
We built two-layer LSTM networks, with each layer containing $256$ LSTM units. 
Dropout was performed at the first and second hidden layers both with a probability of $0.5$.
We used the ADAM updating function \cite{Kingma:2014us}, with a learning rate of $0.001$.
Extending previous work where random starting points have been used for each epoch \cite{Hammerla2016}, during training we dynamically determined mini-batch size $\hat{B}$ and frame lengths $\hat{L}$ randomly within a given range (see Tab.\ \ref{table:mbsz_winLen}).
During test we performed inference at sample level, i.e., both $\hat{L}$ and $\hat{B}$ are set to $1$.
We implemented our framework using Lasagne, a lightweight Theano library to build deep neural networks.
Models were trained on an NVIDIA GPU Tesla K40, and average training times were $9-12$s per epoch for Opportunity/PAMAP2 datasets, and $3 \sim  4$ seconds per epoch for the Skoda dataset.  

\begin{table}[tp]
	\centering
	\tbl{Configuration of mini-batch size and frame length for model training / test.}{
	\begin{tabular}{|c||c|c|c|}
		\hline
		  --			 				& mini-batch size $\hat{B}$  & frame length $\hat{L}$ 	  		\\
		\hline
		training					& \textit{U}(128, 256) 	&  \textit{U}(16, 32)   \\ 
		(mini-batch wise) & & \\
		\hline
		test					& $ 1$ 	&  $1$   \\ 
		(sample wise) & & \\
		\hline
	\end{tabular}}
	\begin{tabnote}
	{\raggedleft
	 $U(\cdot, \cdot)$ denotes a random number generator that draws integers from a discrete uniform distribution with range $(\cdot, \cdot)$.
	 }
	\end{tabnote}	
	\label{table:mbsz_winLen}
\end{table}

\subsubsection{Evaluation}
\label{sec:experiments:models:evaluation}
Overall we are interested in evaluating the robustness of our proposed method for real-world application scenarios.
As such we carefully configured the evaluation protocol aiming for meaningful, practically relevant, and comparable evaluation results.
Based on standard benchmark datasets as described below we adopted evaluation scenarios as they have been used before, which makes our results comparable to related work.
Specifically, we avoided commonly used but unrealistically over-optimistic evaluation protocols such as unconstrained \textit{n}-fold cross-validation \cite{Hammerla:2015vj}.
In accordance with previous work we used hold-out validation and testing protocols where fixed, disjoint portions of a datasets are used for training, validation and testing.
In line with standard procedures in the field, results are reported as mean $F1$-scores:
\begin{equation}\label{eq:Fm}
\bar{F_1} = \frac{1}{K}\sum_{k=1}^K \frac{2 \mathbf{TP}_{k}}{2 \mathbf{TP}_{k} + \mathbf{FP}_{k} + \mathbf{FN}_{k}},
\end{equation}
with $k= 1 \ldots K$ classes (activities) considered.
$\mathbf{TP}_k, \mathbf{FP}_k, \mathbf{FN}_k$ denote the number of true positive, false positive, and false negative predictions, respectively.

{\color{black}Through our experimental evaluation we analyse the effects the parameters (details explained below) of our approach have on the overall recognition performance. For every model configuration we ran $30$ repetitions of the particular experiments and report averaged F1 scores for each set of experiments (plus standard deviation). Given the elements of --intentional-- randomness in our Ensemble creation procedure, this protocol captures the performance as it is to be expected for real-world deployments and enables formal statistical significance testing. For the latter we employ two-tailed independent $t$-tests and report significance levels through $p$ values, where $p \leq 0.05$, $p \leq 0.01$ and $p \leq 0.001$ correspond to *, ** and ***, respectively (in line with the literature, e.g., \cite{morales2016deep}).}

Note that we report sample-wise prediction results, which is according to our overall motivation and in line with previous work \cite{Hammerla2016} but different to frame-wise results as, for example, reported in \cite{Ordonez2016}.
For comparability we reran experiments from the literature and where necessary converted frame-based results to sample-wise prediction.

\subsection{Datasets}
\label{sec:experiments:datasets}
In line with related work in this field 
we evaluated our system on three benchmark datasets that have widely been used within the community. 
These datasets contain continuous sensor readings from inertial measurement units that were worn by participants of the particular studies at different positions at their bodies.
The chosen datasets correspond to three diverse but at the same time typical tasks for human activity recognition in the field. 
{\color{black} In line with our general ambition to develop recognition systems for real-world HAR applications, our primary concern is related to the evaluation of challenging, realistic benchmark datasets -- most prominently Opportunity, which, arguably, represents one of the most difficult datasets to date. However, in order to avoid potential overfitting to one specific scenario and --more importantly-- to demonstrate the generalisability of the developed method, we evaluate our approach on datasets that might not be directly mirroring real-world deployments but still are widely used in the research community (e.g., PAMAP2).} 

\begin{figure}[tp]
	\hspace*{-2em}
	\subfloat[Opportunity]{
	      \includegraphics[height=3cm]{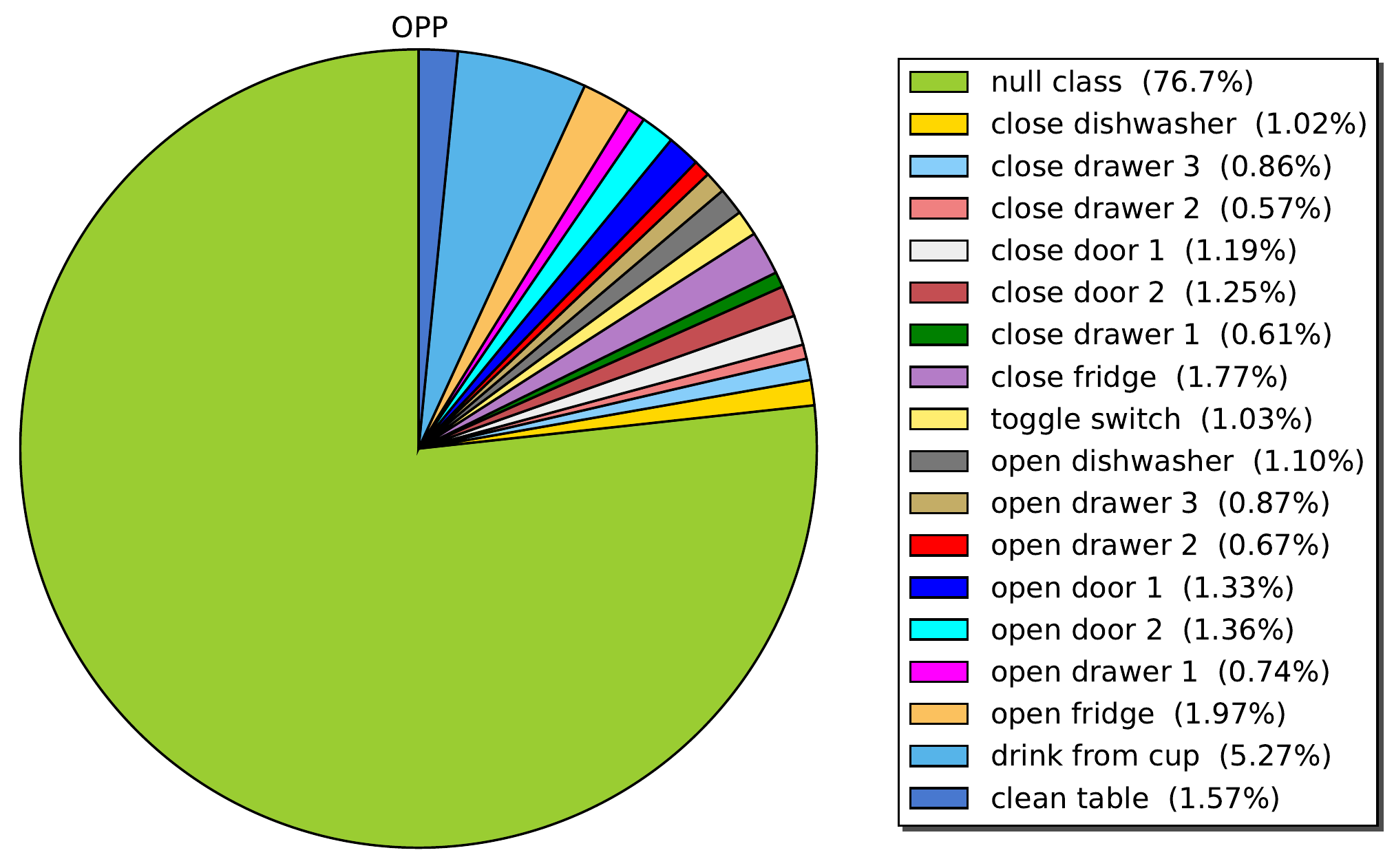}
	      \label{fig:datadistr:opportunity}
	}
	\subfloat[PAMAP2]{
	      \includegraphics[height=3cm,trim={2.1cm 0 0 0},clip]{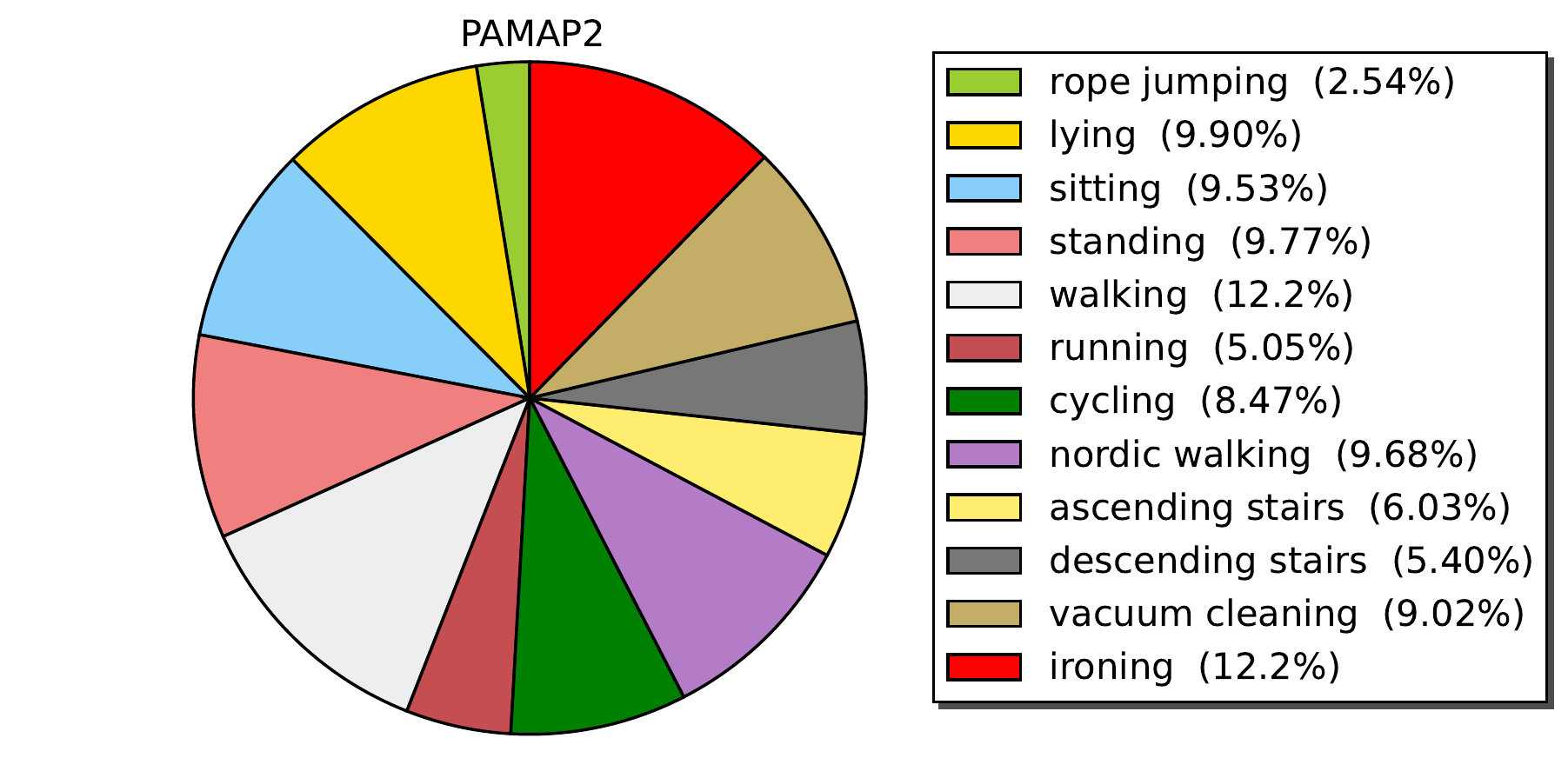}
	      \label{fig:datadistr:pamap2}
	}
	\hspace*{-1em}
	\subfloat[Skoda]{
	      \includegraphics[height=3cm,trim={2cm 0 0 0},clip]{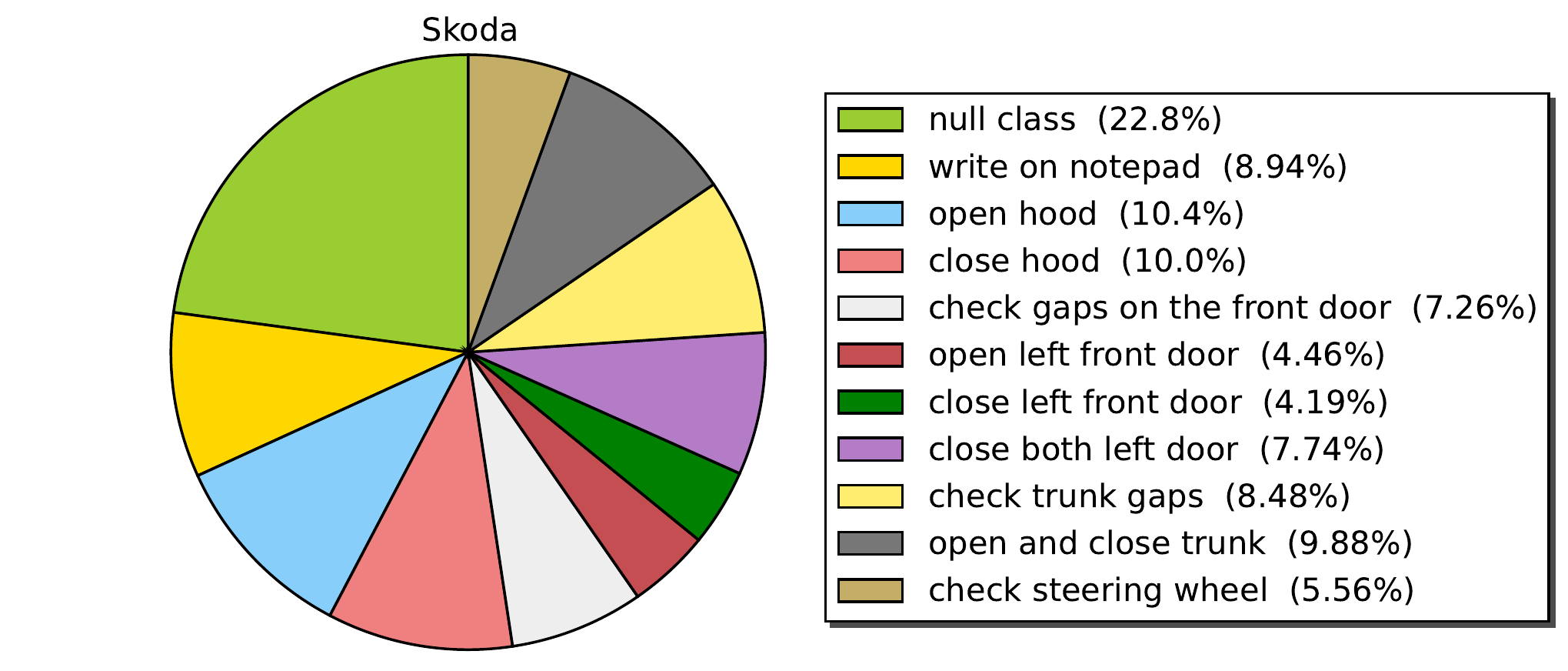}
	      \label{fig:datadistr:skoda}
	}
	\vspace*{-0.5em}
      \caption{Class distributions of the three datasets used for experimental evaluation (best viewed in colour).}
      \label{fig:datadistr}
\end{figure}

\begin{figure}[tp]
	\vspace*{-1em}
	\centering
	\subfloat[Opportunity]{
		\includegraphics[width=\textwidth,trim={0 0 0 2.5cm},clip]{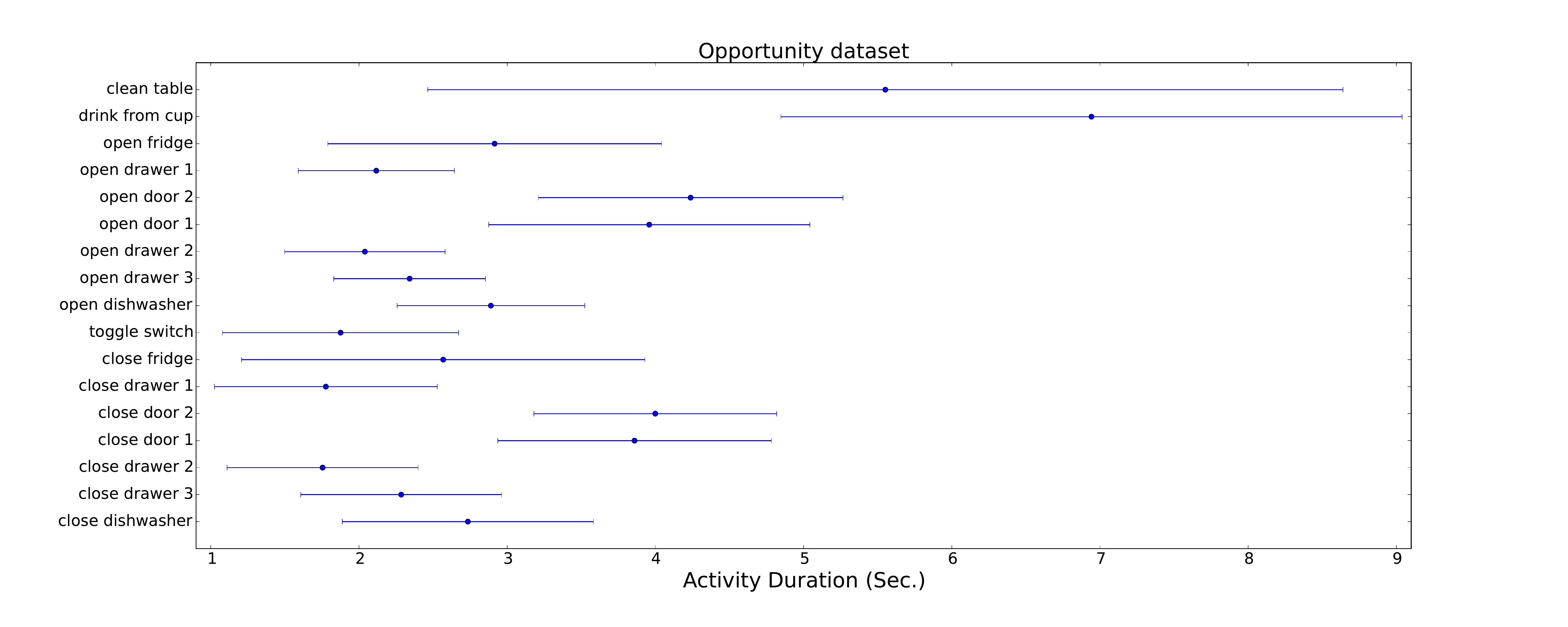}
		\label{fig:durations:opportunity}
	}
	\\
	\subfloat[PAMAP2]{
		\includegraphics[width=\textwidth,trim={0 0 0 2cm},clip]{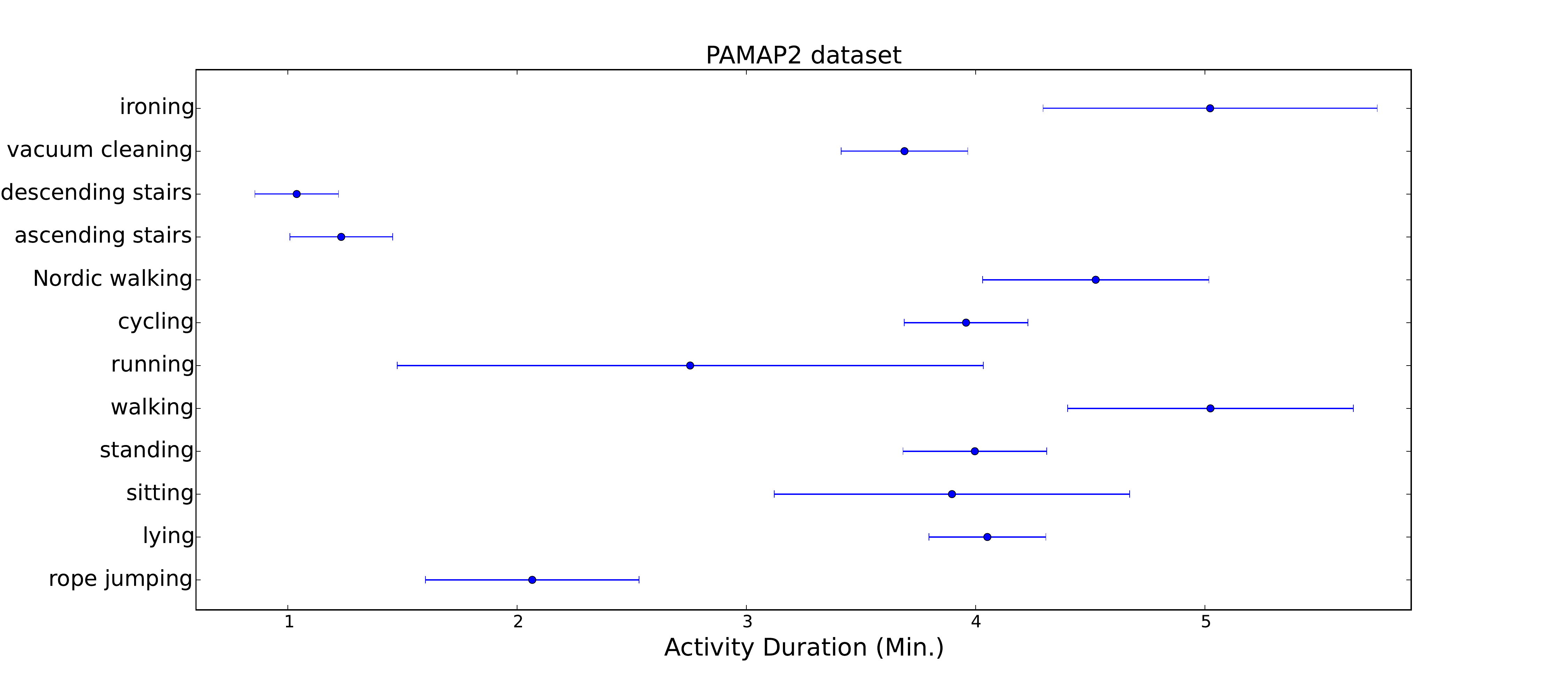}
		\label{fig:durations:pamap2}
	}
	\\
	\subfloat[Skoda]{
		\includegraphics[width=\textwidth,trim={0 0 0 1.1cm},clip]{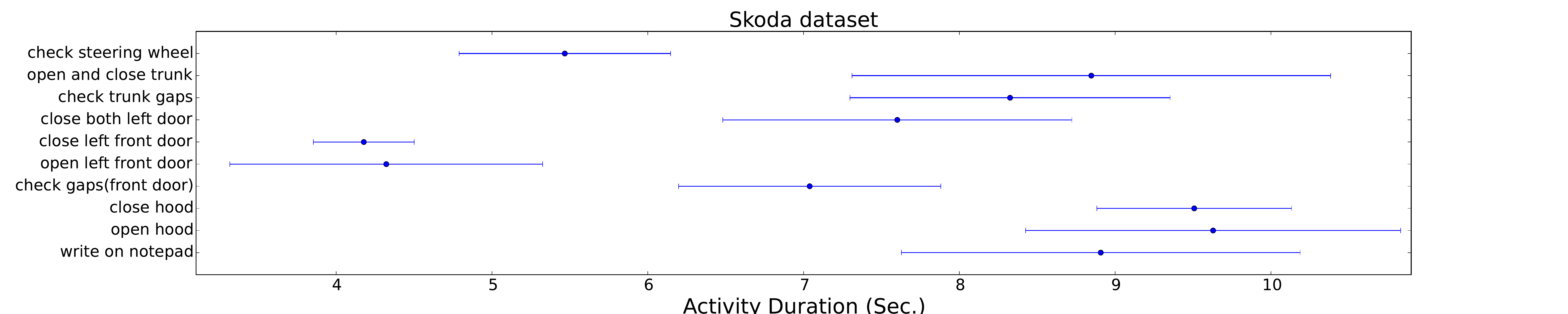}
		\label{fig:durations:skoda}
	}
	\caption{{\color{black}Average durations of activities for the three benchmark datasets used [mean $\pm$ 	std]}}
	\label{fig:durations}
\end{figure}

Fig.\ \ref{fig:datadistr} shows the class distributions for the three datasets illustrating one of the aforementioned challenges to HAR in ubiquitous and wearable computing.
{\color{black}In addition, Fig.\ \ref{fig:durations} gives an overview of average durations of activities of interests in all three datasets, which shows the substantial variability and thus further challenges for automated recognition.}
Especially for the Opportunity dataset a substantially imbalanced class distribution with an additional strong bias towards the NULL class can be observed (Fig.\ \ref{fig:datadistr:opportunity}).
This is not unusual for mobile application scenarios where activities of interest are often underrepresented in the overall dataset due to the continuous recording in real-life scenarios.
Prominent examples for such scenarios are health assessments where problematic activities or behaviours are typically the rare exception within other, less problematic activities.
For example, freeze of gait in Parkinson's is a common problem amongst patients but throughout the day such freezing episodes are rare -- though very critical w.r.t.\ health and wellbeing of the patient and thus require reliable recognition (e.g., \cite{bachlin2010wearable}).
Another related example from the literature is the automated assessment of problem behaviour in individuals with Autism, where aggressive behaviours do occur frequently yet they typically represent the minority of everyday activities thus resulting in imbalanced datasets \cite{Ploetz:2012vm}.
PAMAP2 is a dataset that was recorded in a much more rigorously scripted and thus constrained manner, which results in a well balanced class distribution -- mainly reasoned by the absence of a NULL class (Fig.\ \ref{fig:datadistr:pamap2}).
Finally, the Skoda dataset \cite{Stiefmeier2008-WAT} shows a reasonably well balanced class distribution for the activities of interest (Fig.\ \ref{fig:datadistr:skoda}).
However, overall the dataset is still dominated by the NULL class rendering it biased as one would expect for real-world scenarios as discussed before.
{\color{black}Note that for most real-world deployments of HAR it is rather unrealistic to assume that simple pre-processing techniques would allow to effectively eliminate background data (NULL class) from continuous recordings of sensor data. The reason for this is that background activities often are very diverse and the classification into background vs activities of interest typically corresponds to a non-trivial task itself, which typically requires more than simple filtering. Furthermore, whilst it may be straightforward to identify and eliminate faulty sensor readings, noisy sensor data, that are very common in mobile scenarios, represent a different kind of challenge that require more sophisticated measures to tackle.}
In what follows we will provide detailed summaries of the three datasets and how they have been used in our experimental evaluations.

\subsubsection{Opportunity}
\label{sec:experiments:datasets:opportunity}
The Opportunity dataset is the result of a concerted effort towards collecting a benchmark dataset for human activity recognition using body-worn sensors  \cite{Chavarriaga2013a}.
It consists of annotated recordings from a total of four participants who wore an armada of sensors and were instructed to carry out everyday life domestic activities, specifically focusing on kitchen routine. 
Data was recorded using Inertial Measurement Units attached at $12$ on-body positions.
The sampling frequency for all IMUs was $30$Hz, and annotations are provided for $18$ mid-level activities, such as Open Door / Close Door. 
Every participant performed five different runs of the kitchen activities.
We replicate the training, validation, and test protocol from \cite{Hammerla2016}.
Accelerometer recordings from the upper limbs, the back, and complete IMU data from both feet were used resulting in a $79$-dimensional dataset. 
All data were normalised to zero mean and unit variance.
The second run from participant $1$ was used as validation set, and 
we use runs $4$ and $5$ from participants $2$ and $3$ as test. 
The remaining data were used for training (total of approx.\ $650$k samples).

\subsubsection{PAMAP2}
\label{sec:experiments:datasets:pamap2}
This dataset was recorded in a scripted and thus rather constrained setting where nine participants were instructed to carry out a total of $12$ activities of daily living, covering domestic activities and various sportive exercises (Nordic walking, running, etc) \cite{Reiss2012}.
Full IMU data (accelerometer, gyroscope, magnetometer) plus temperature and heart rate data were recorded from body-worn sensing platforms that were attached to the hand, chest and ankle.
A total of more than $10$ hours of data was collected. 
The resulting dataset has $52$ dimensions. 
Sensor data were down-sampled to $33$Hz to match the temporal resolution of the other datasets used. 
All samples were normalised to zero mean and unit variance.
Again replicating the protocol from \cite{Hammerla2016}, we used runs $1$ and $2$ for participant $5$ for validation, and runs $1$ and $2$ from the sixth participant as test. 
The remaining data was used for training (total of approx.\ $473$k samples). 

\subsubsection{Skoda}
\label{sec:experiments:datasets:skoda}
The third dataset used for our experimental evaluation has been recorded in a manufacturing scenario.
Specifically, it covers the problem of recognising activities of assembly-line workers in a car production environment \cite{Stiefmeier2008-WAT}. 
In the study that led to the collection of the Skoda dataset a worker wore a number of accelerometers while undertaking manual quality checks for correct assembly of parts in newly constructed cars.
These checks translate into $10$ manipulative gestures of interest, including checking the boot, opening/closing engine bonnet, boot and doors, and turning the steering wheel.
Accelerometer data were down-sampled to $33$Hz, and we normalised them to zero mean and unit variance.
We used data (raw and calibrated, according to the dataset description in the original publication) recorded using the $10$ accelerometers worn on the right arm of the worker, resulting in $60$-dimensional input data.
As for the other two datasets we use hold-out validation and test protocols.
The training set contains the first $80$\% of each class (a total of approx.\ $190$k samples), validation the next $10$\%, and test contains the remainder of the dataset.

\subsection{Results}
\label{sec:experiments:results}





\begin{figure}[tp]
	\vspace*{1em}
	\begin{tabular}{ccc}
	 \hspace*{-0.2cm}	
	 \includegraphics[width=0.37\textwidth,trim={0 0 0 1cm},clip]{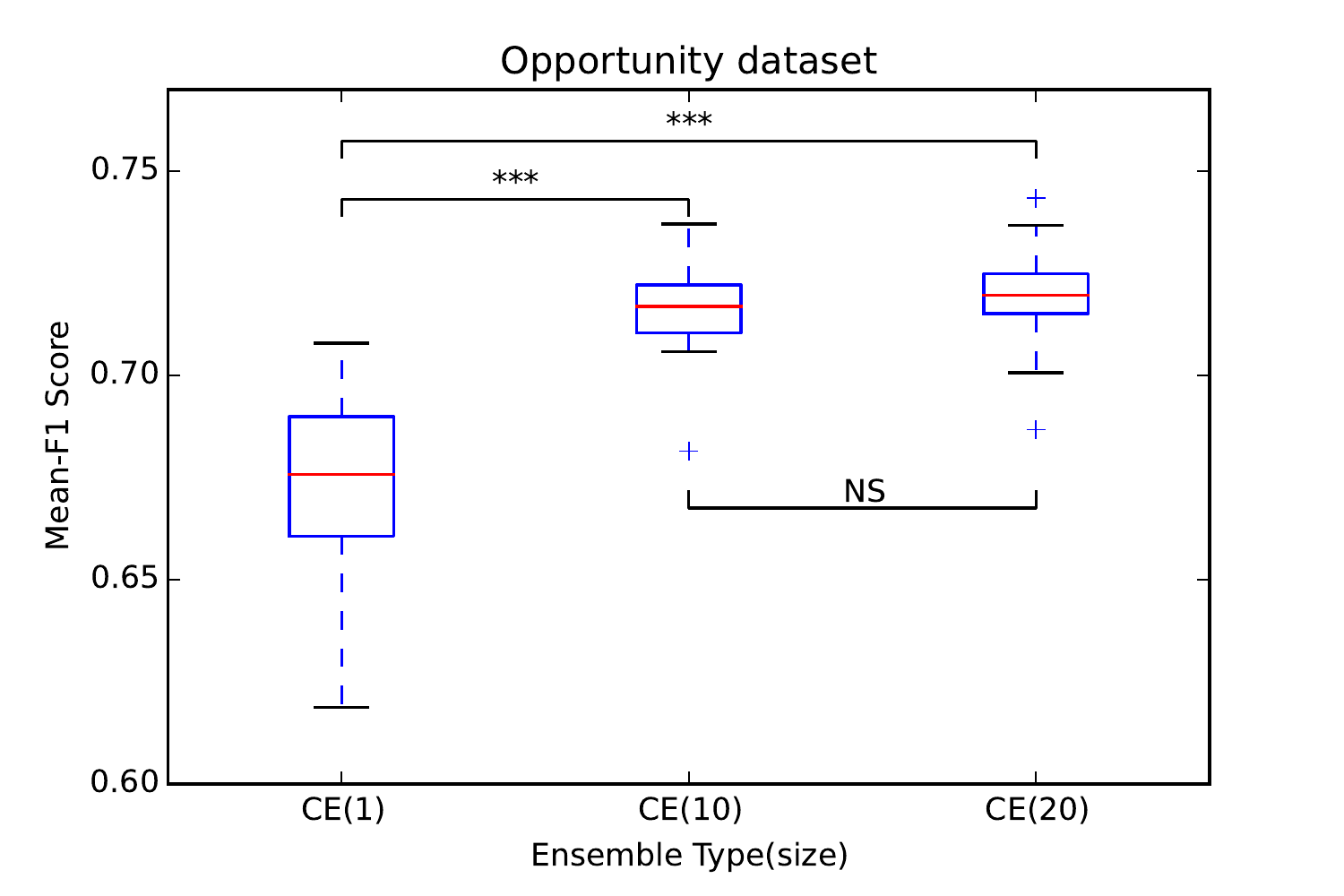}
	 &
	 \hspace*{-1cm}
	 \includegraphics[width=0.37\textwidth,trim={0 0 0 1cm},clip]{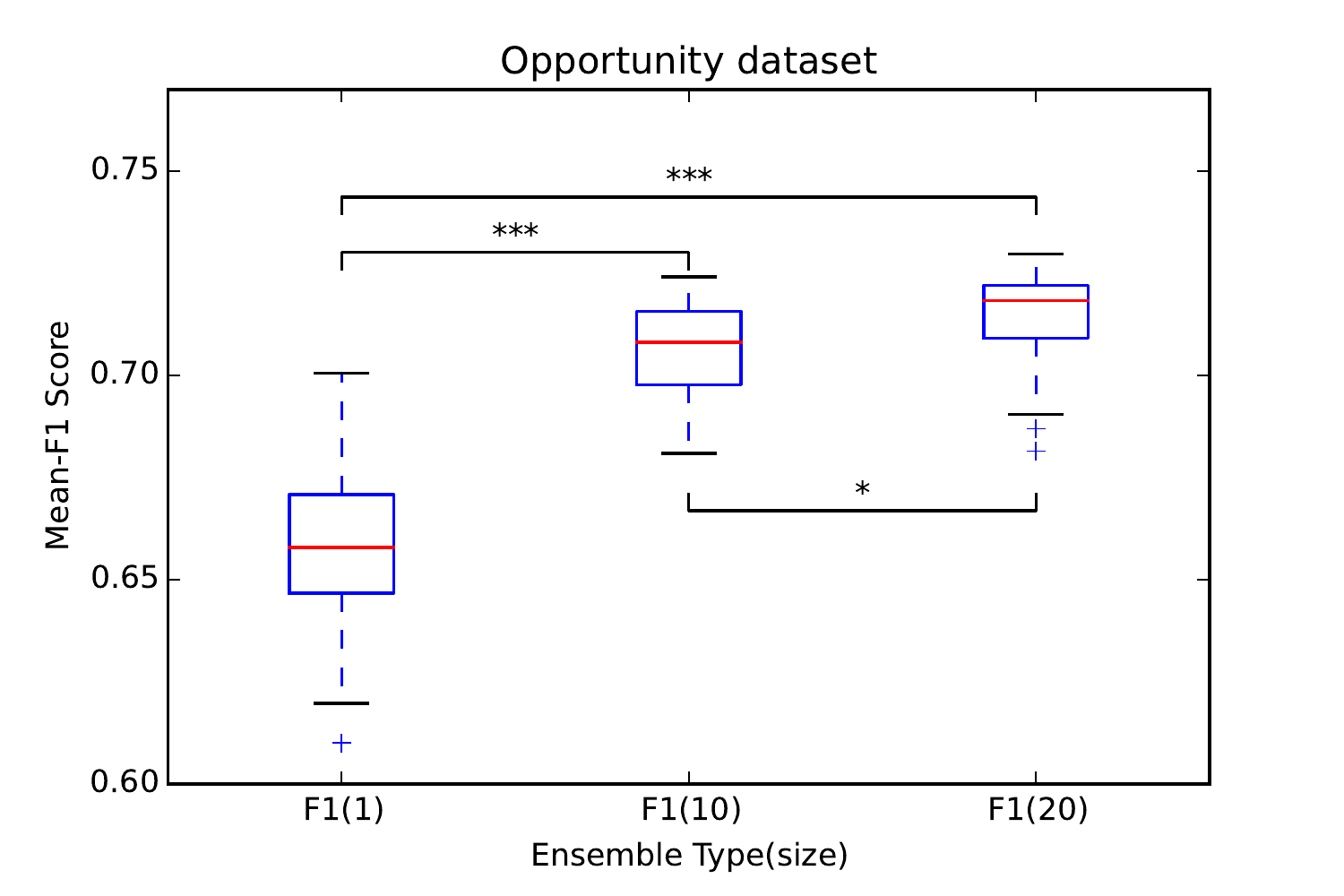}
	 &
	 \hspace*{-1cm}
      	\includegraphics[width=0.37\textwidth,trim={0 0 0 1cm},clip]{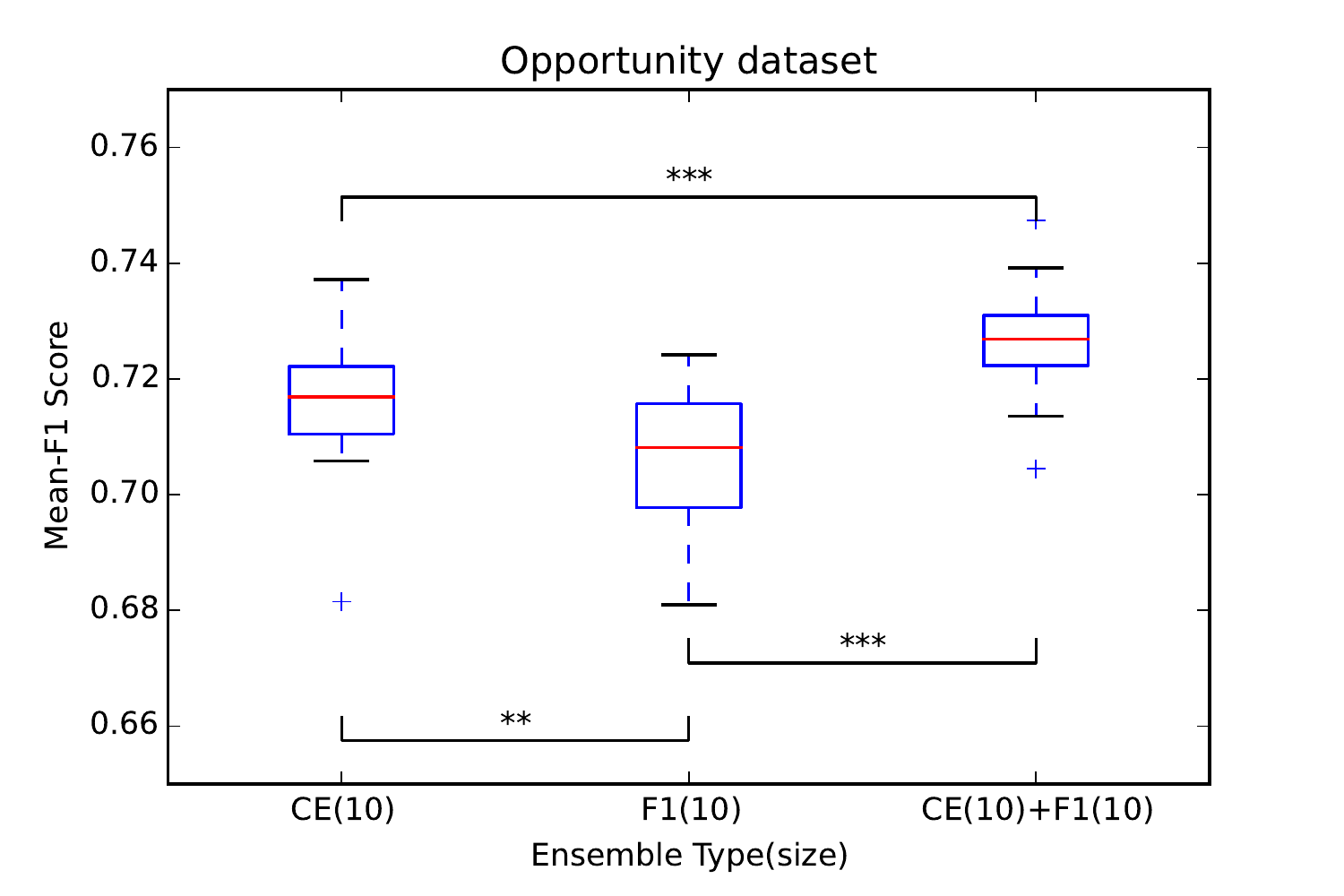}
	\\
	\multicolumn{3}{c}{Opportunity}\\
	\\
	 \hspace*{-0.2cm}	
	\includegraphics[width=0.37\textwidth,trim={0 0 0 1cm},clip]{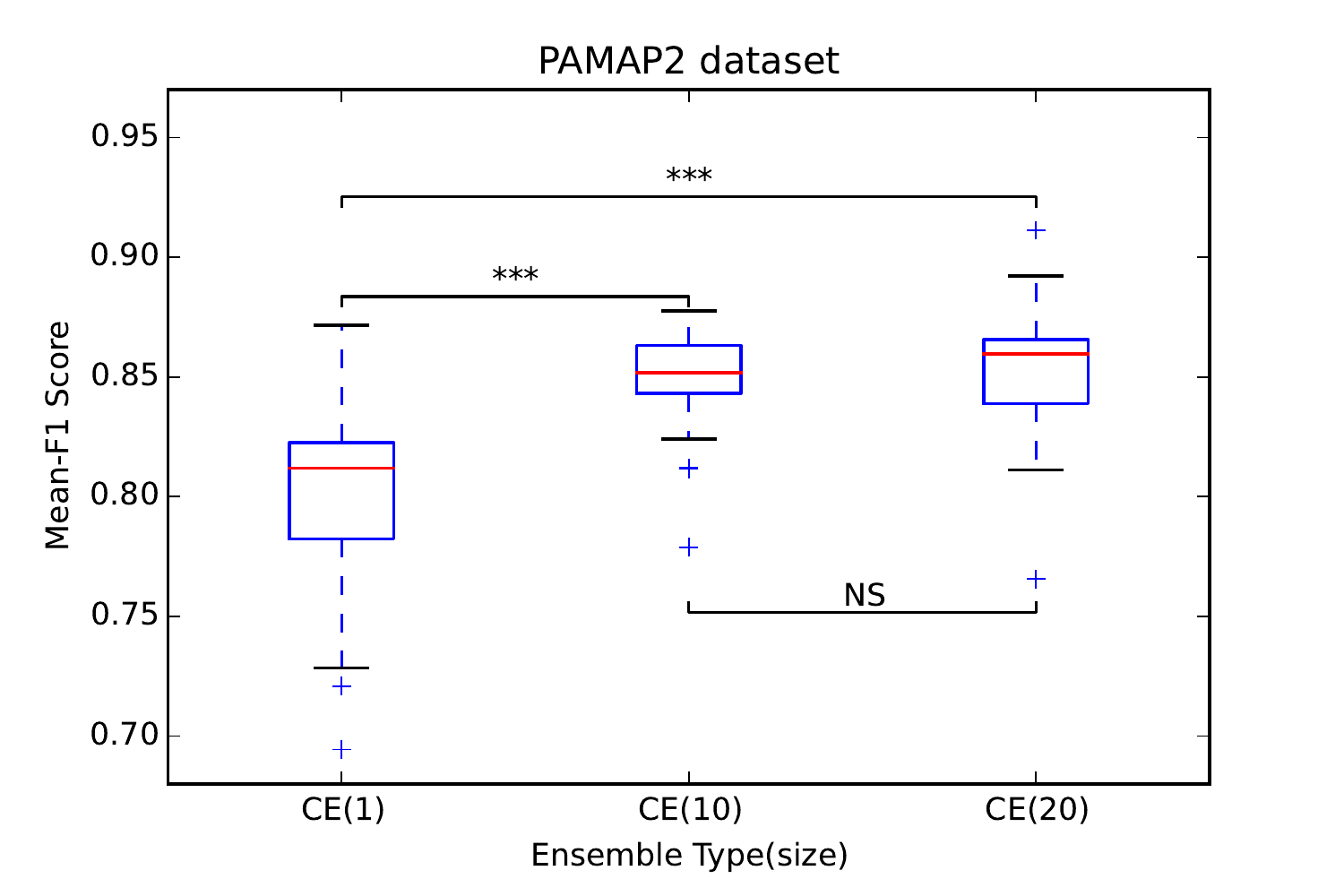}
	&
	 \hspace*{-1cm}	
      	\includegraphics[width=0.37\textwidth,trim={0 0 0 1cm},clip]{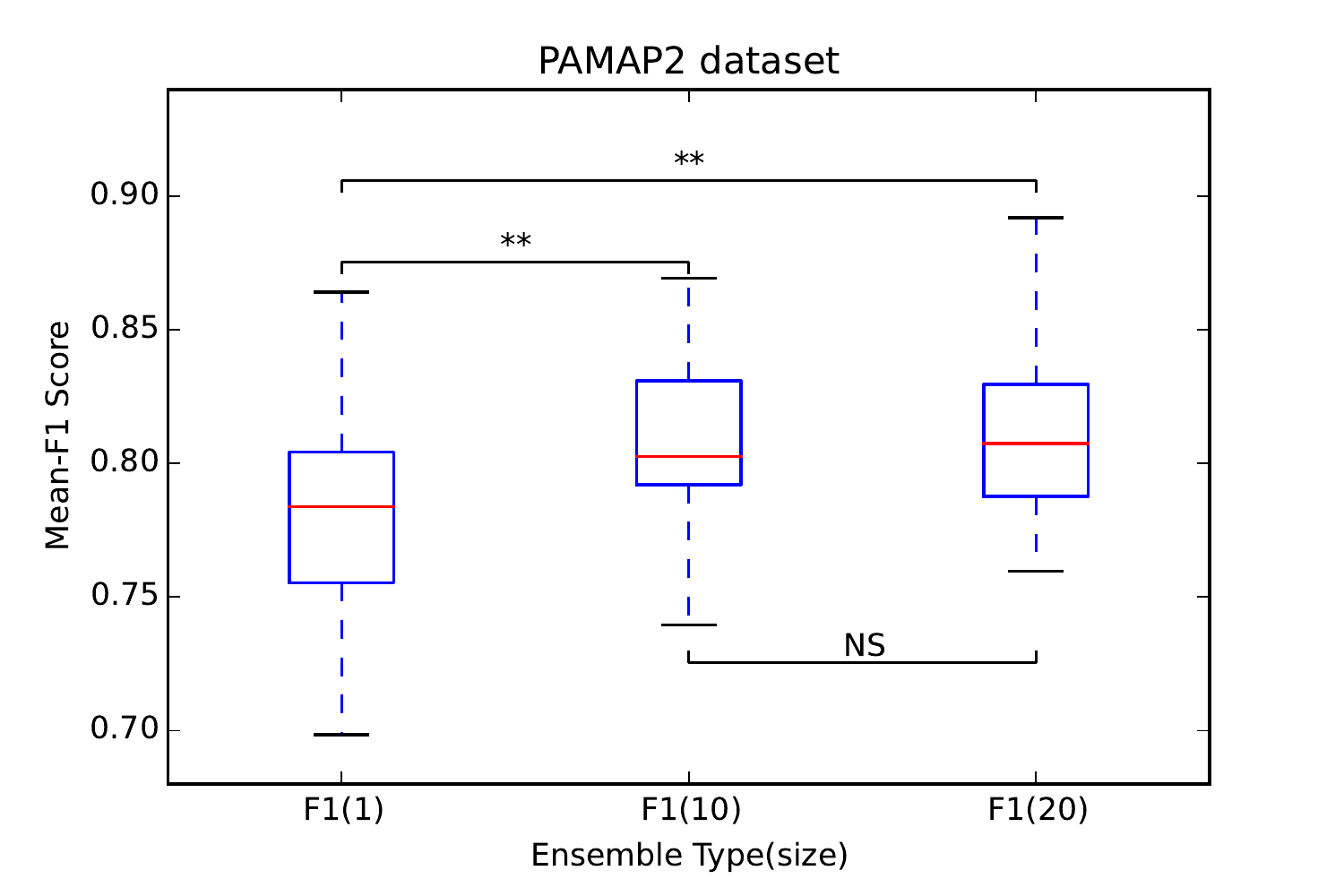}
	&
	 \hspace*{-1cm}	
	\includegraphics[width=0.37\textwidth,trim={0 0 0 1cm},clip]{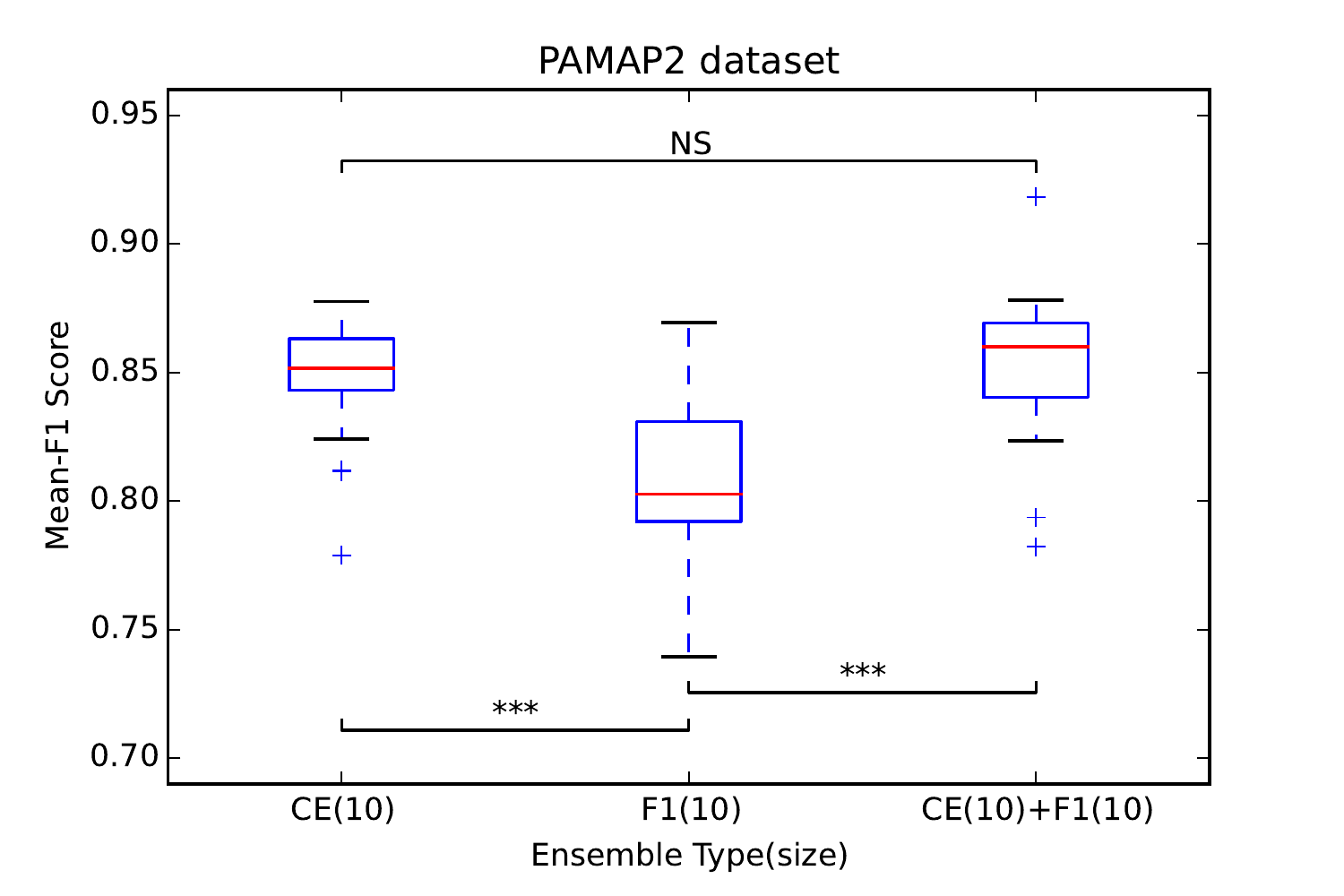}
	\\
	\multicolumn{3}{c}{PAMAP2}\\
	\\
	 \hspace*{-0.2cm}	
	\includegraphics[width=0.37\textwidth,trim={0 0 0 1cm},clip]{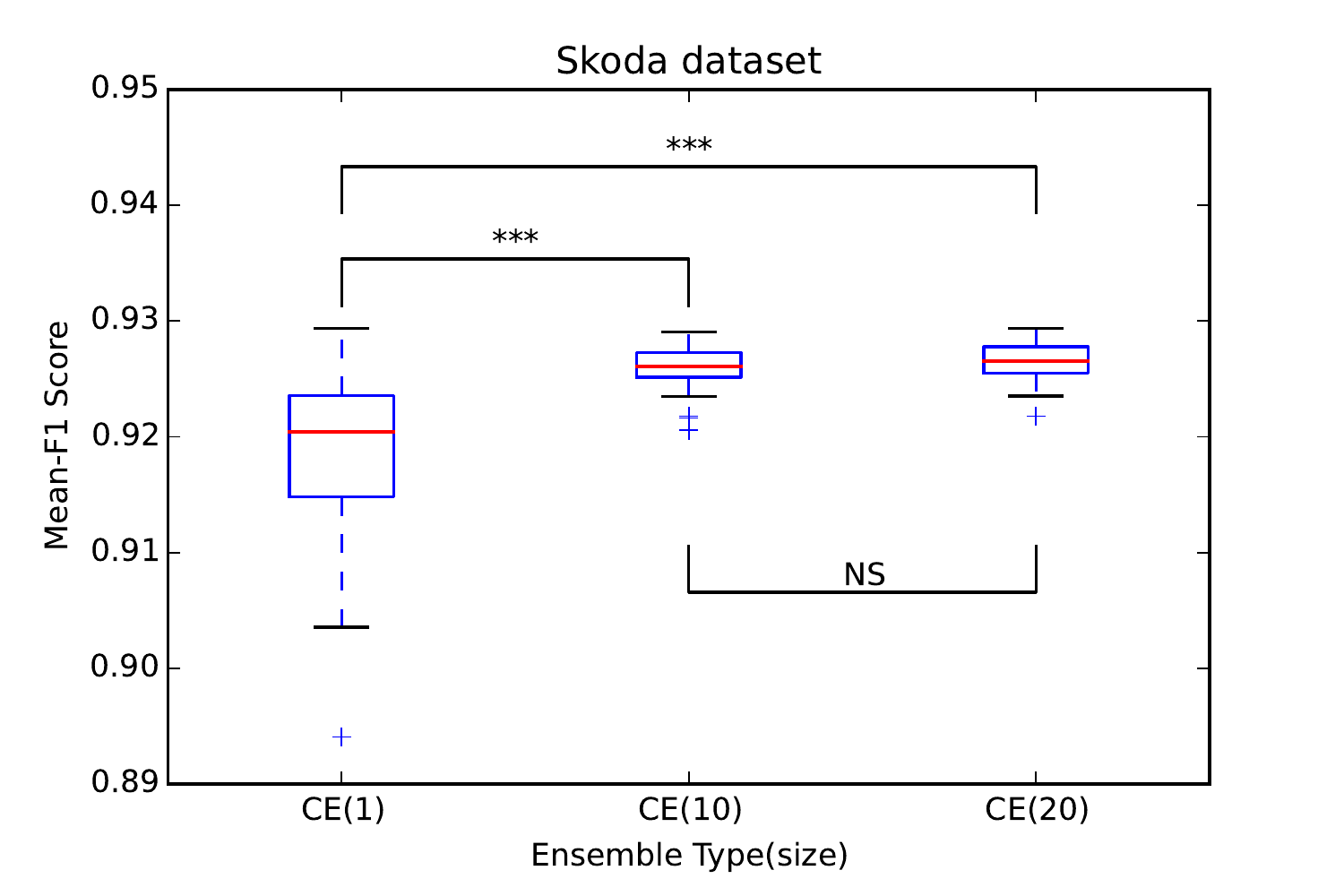}
	&
	 \hspace*{-1cm}	
      	\includegraphics[width=0.37\textwidth,trim={0 0 0 1cm},clip]{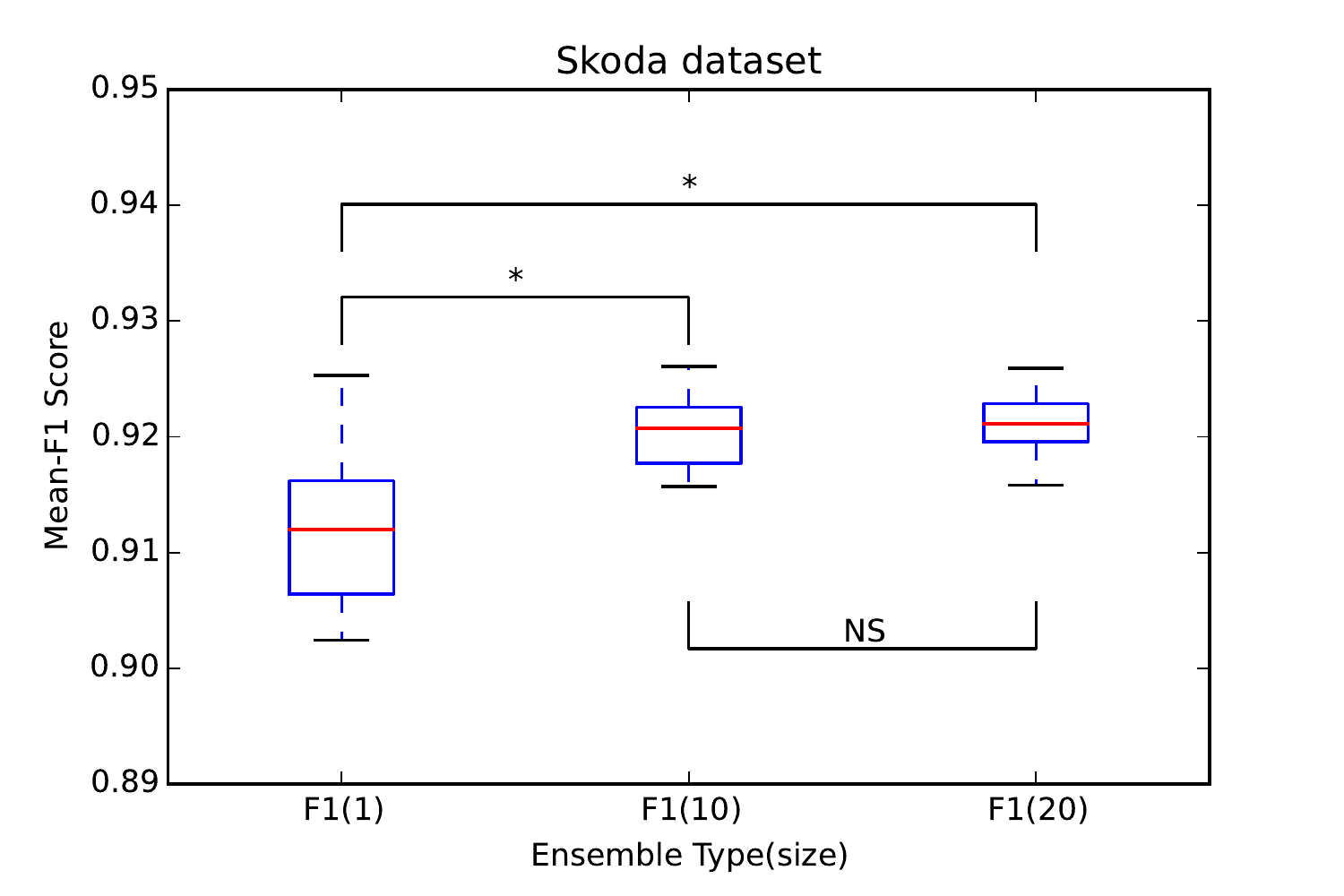}
	&
	 \hspace*{-1cm}	
	\includegraphics[width=0.37\textwidth,trim={0 0 0 1cm},clip]{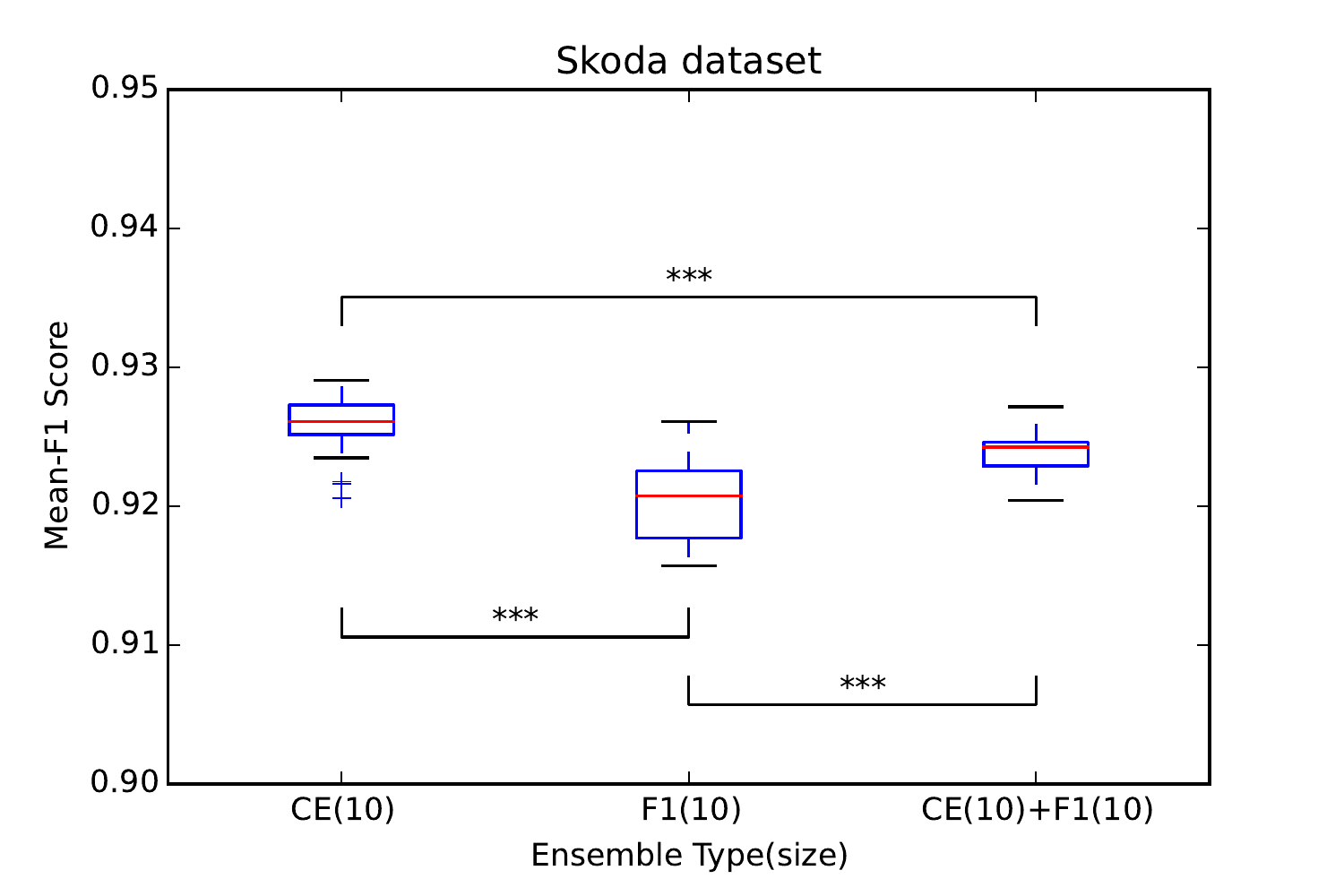}
	\\
	\multicolumn{3}{c}{Skoda}\\
	\end{tabular}
	%
      	\caption{{\color{black}Statistical significance analysis (two-tailed independent $t$-test) of recognition experiments using variants of LSTM ensembles, based on $30$ trials each. See text for description, best viewed in colour.}}
	\label{fig:significance} 
\end{figure}

\begin{figure}[tp]
	\vspace*{-3em}
	\centering
	\subfloat[Opportunity]{
		\includegraphics[width=0.5\textwidth,trim={0 0 0 1cm},clip]{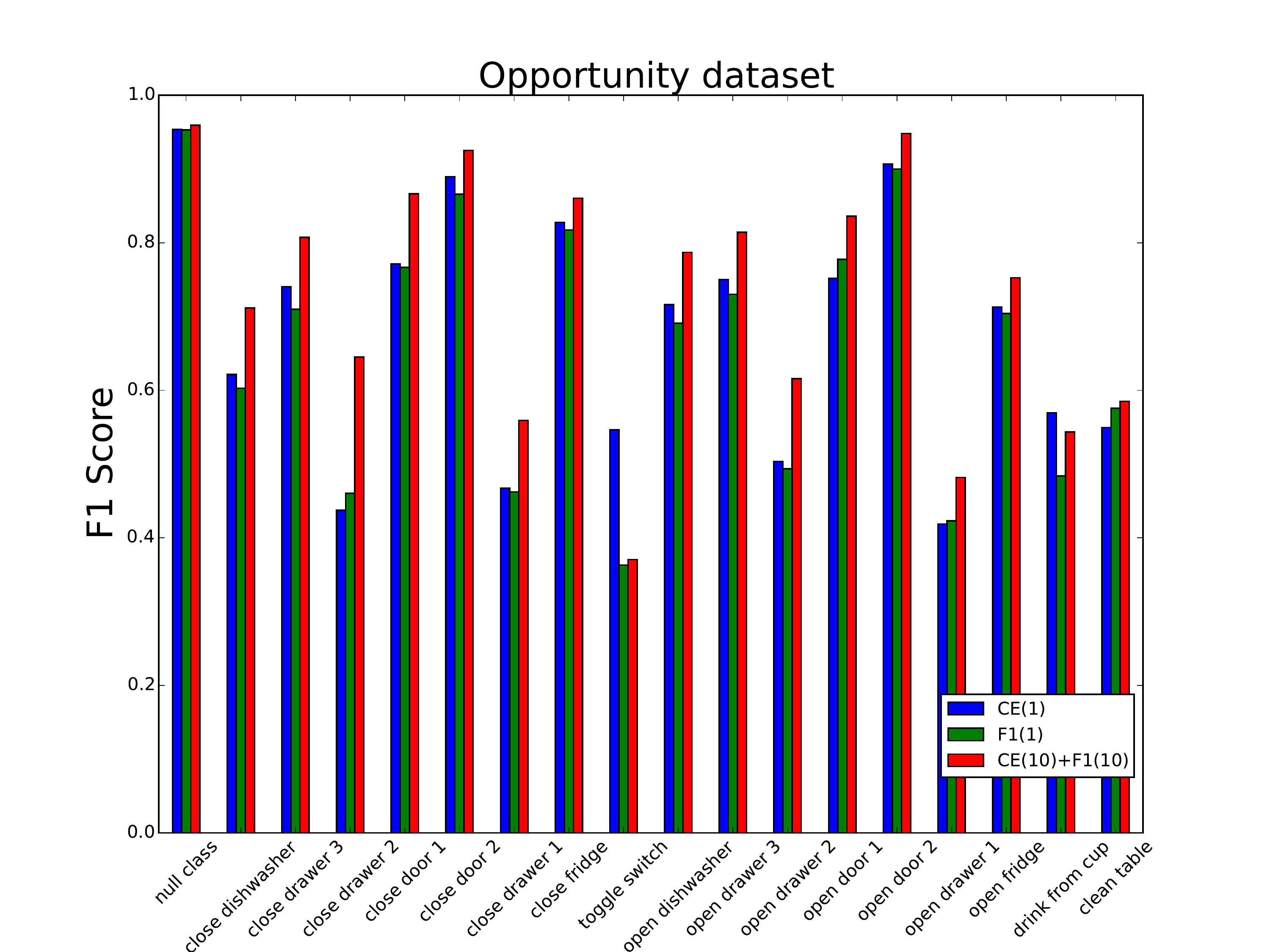}
		\label{fig:results-class:opportunity}
	}
	\subfloat[PAMAP2]{
		\includegraphics[width=0.5\textwidth,trim={0 0 0 1cm},clip]{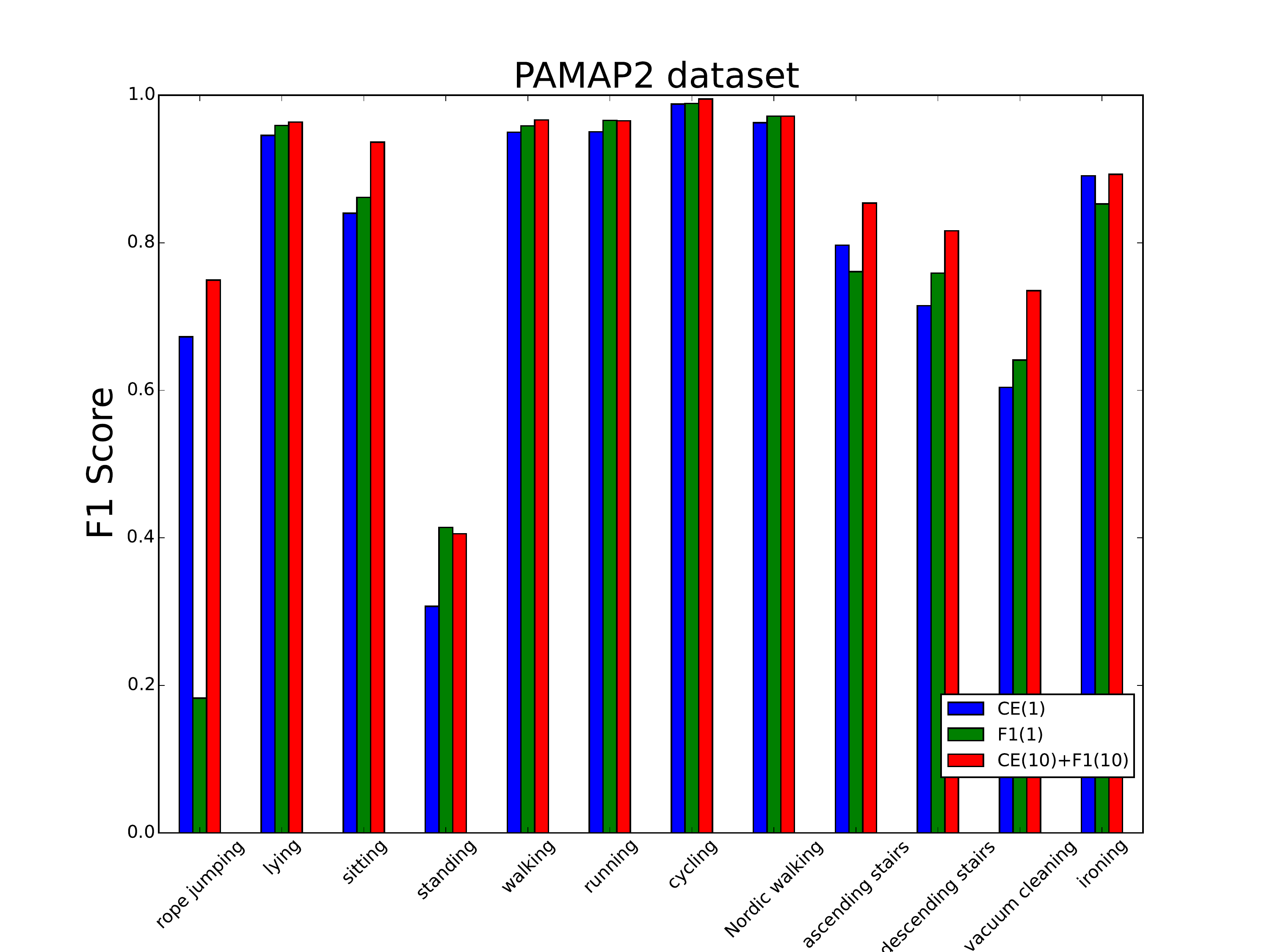}
		\label{fig:results-class:pamap2}
	}
	\\[-0.3em]
	\subfloat[Skoda]{
		\includegraphics[width=0.5\textwidth,trim={0 0 0 1cm},clip]{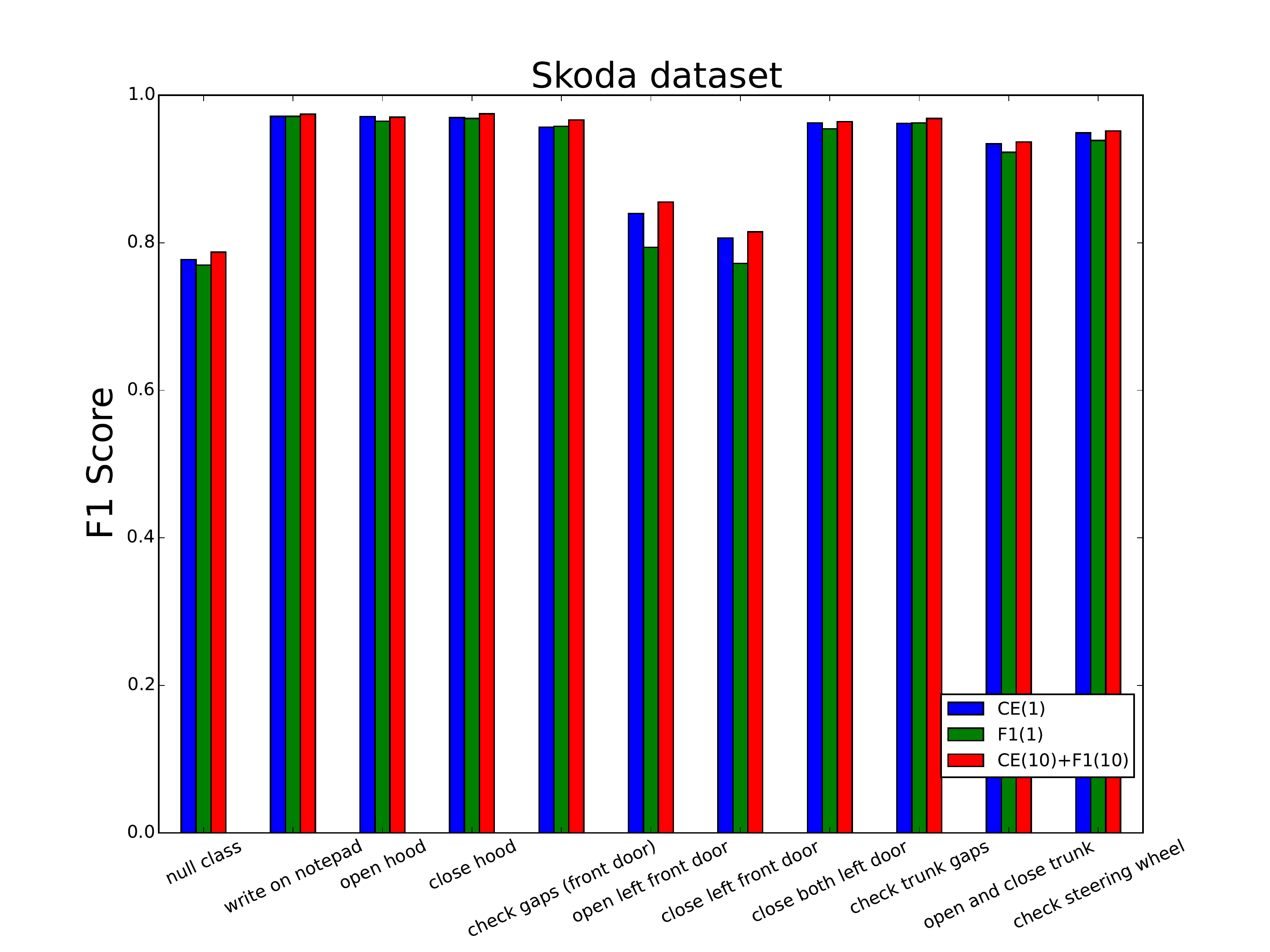}
		\label{fig:results-class:skoda}
	}
	\vspace*{-1em}
	\caption{{\color{black}Class-wise recognition results for experiments on benchmark datasets (F1 scores; averages over $30$ trials each).}}
	\label{fig:results-class}
\end{figure}

\begin{figure}[tp]
	\vspace*{-3em}
	\centering
	\subfloat[Opportunity]{
		\adjincludegraphics[width=0.33\textwidth,trim={0cm 4cm 9cm 5cm},clip]{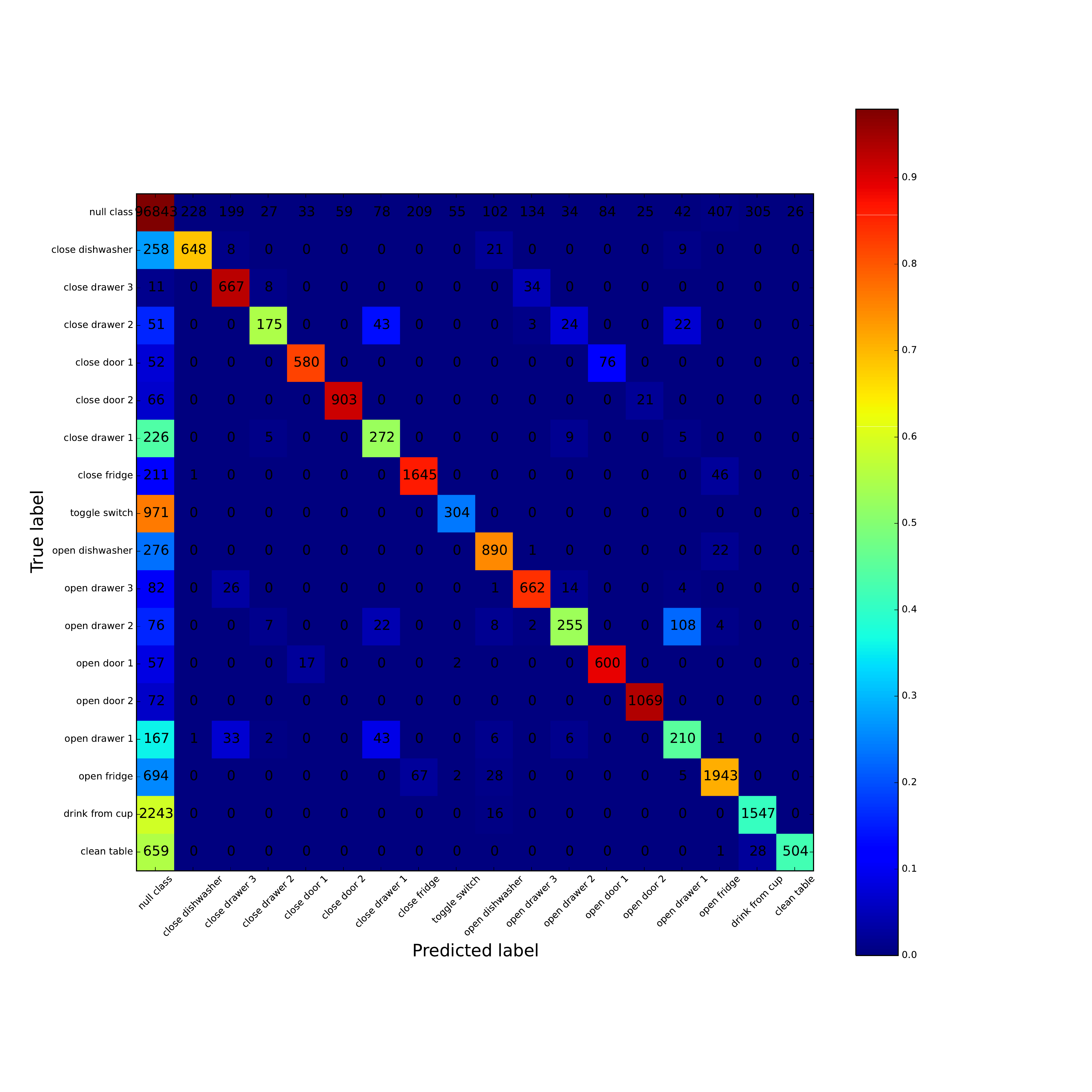}
		\label{fig:results-confusion:opportunity}
		\vspace*{-2em}		
	}
	\hspace*{-0.3cm}
	\subfloat[PAMAP2]{
		\includegraphics[width=0.312\textwidth,trim={1.7cm 4cm 9cm 5cm},clip]{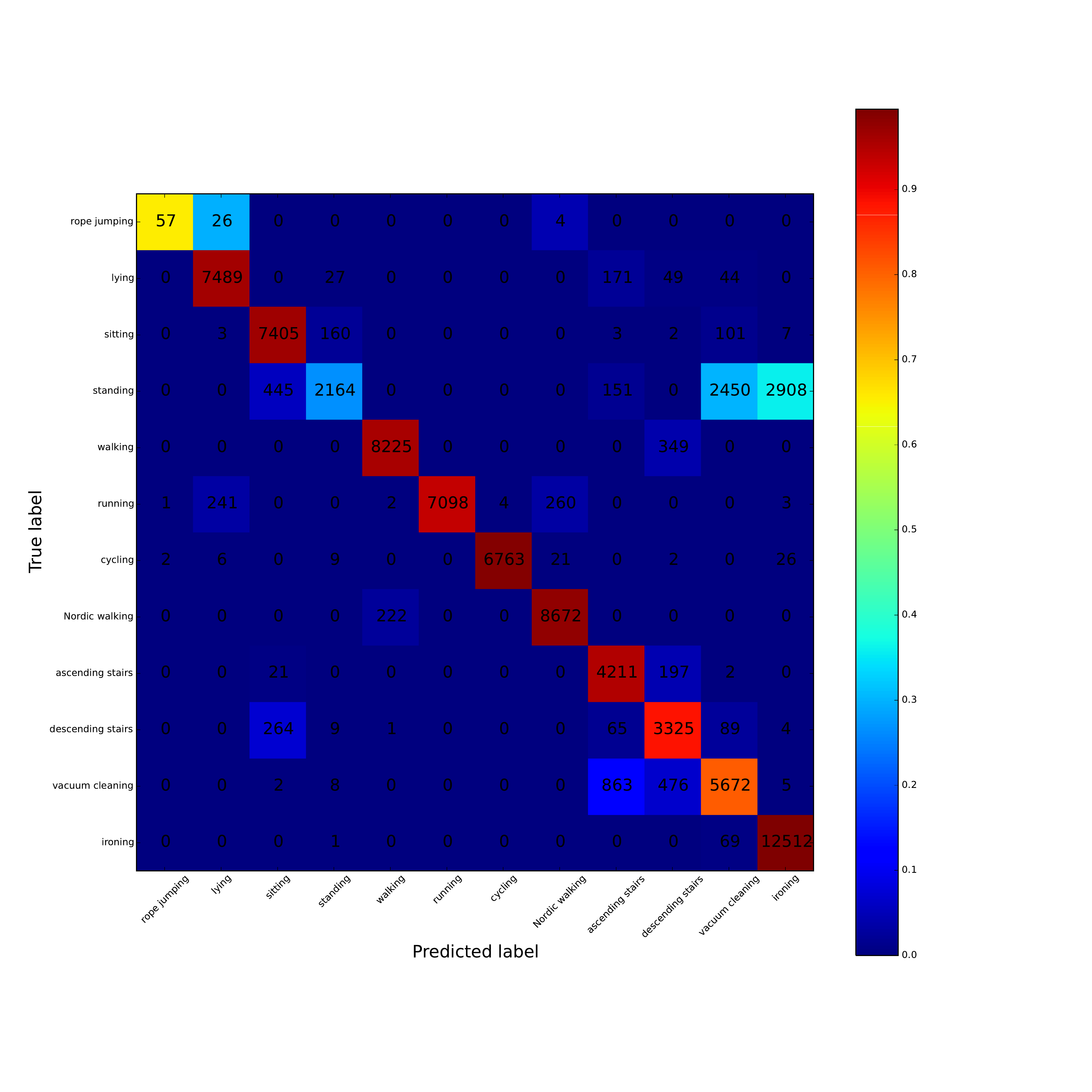}
		\label{fig:results-confusion:pamap2}
		\vspace*{-2em}
	}
	\hspace*{-0.3cm}
	\subfloat[Skoda]{	
		\includegraphics[width=0.3423\textwidth,trim={18 4cm 7cm 5cm},clip]{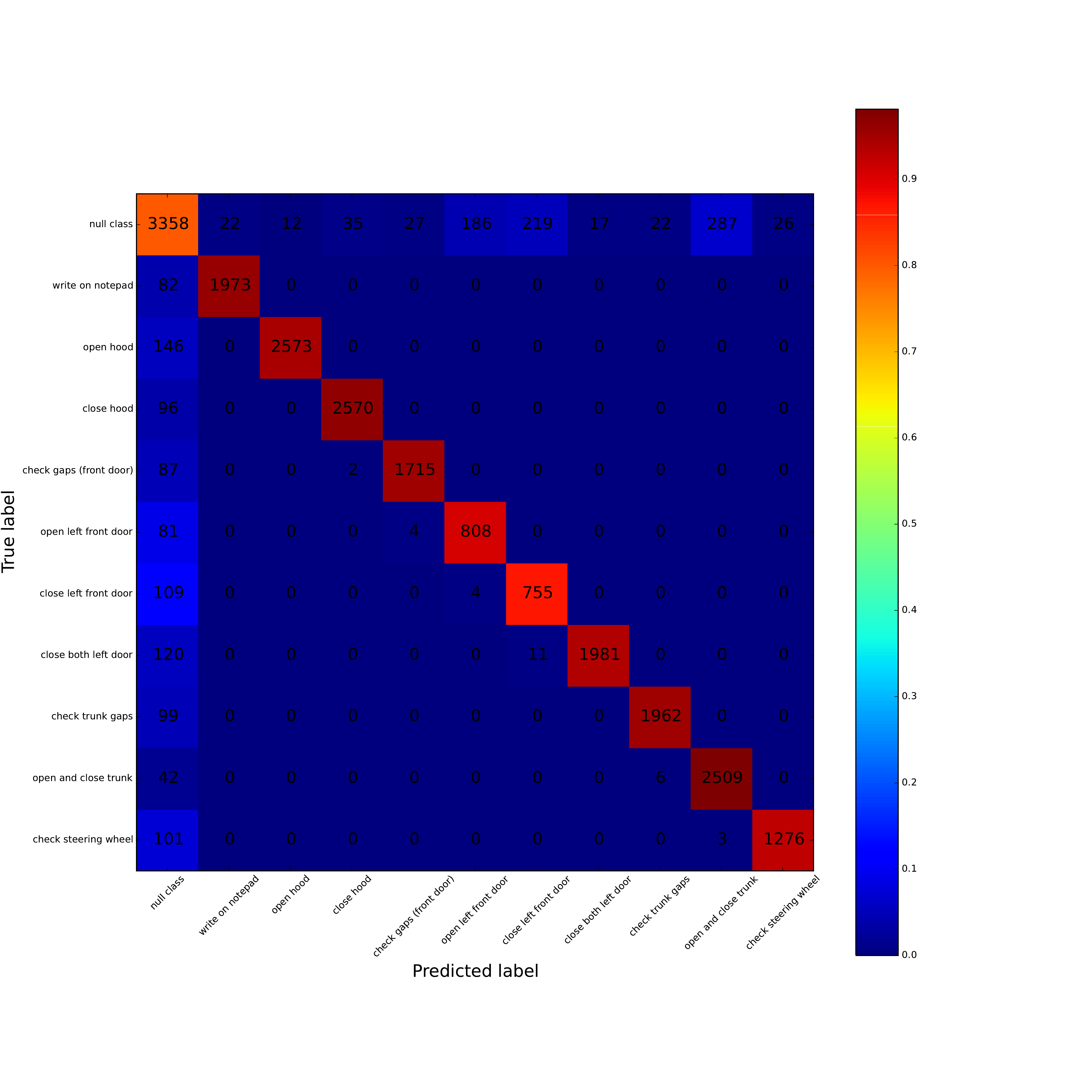}
		\label{fig:results-confusion:skoda}
		\vspace*{-2em}
	}
	\vspace*{-1em}
	\caption{{\color{black}Confusion matrices for all recognition experiments (Ensemble model (CE(10)+F1(10)),  best viewed in colour).}}
	\label{fig:results-confusion}
\end{figure}

Tab.\ \ref{tab:results} provides an overview of the results of our experimental evaluation based on Ensembles of LSTM learners as achieved for the three benchmark datasets.
The table contains results {\color{black}(mean $F1$ scores and standard deviations based on $30$ repetitions each, cf.\ Sec.\ \ref{sec:experiments:models:evaluation} for explanation)} for various model configurations {\color{black}(specified in first column)}:

\begin{description}
	\item [Number of base learners for Ensembles]
	We analysed the effect the incorporation of LSTM models into the overall recognition framework has on the total HAR performance.
	For every dataset we tested Ensembles with the best $M=1$ ('Single'), $M=10$, and $M=20$ individual LSTM learners.

	\item [Loss functions]
	We evaluated the use of standard cross entropy --'Ensemble (CE)'-- and F1 scores --'Ensemble (F1))'-- as loss functions, as well as the combination of models based on either of these loss functions ('Ensemble (CE+F1)').
\end{description}
%
%
{\color{black}Fig.\ \ref{fig:significance} illustrates results of statistical significance tests for all experiments as described in Sec.\ \ref{sec:experiments:models:evaluation}. Furthermore, Figs.\ \ref{fig:results-class} and \ref{fig:results-confusion} show class-specific recognition results and confusion matrices for all three recognition tasks. Averaged confusion matrices in Fig.\ \ref{fig:results-confusion} are given for the best model configuration, namely Ensembles with combined CE+F1 loss function and $10$ base learners each.}

{\color{black} The results unveil a number of interesting effects our Ensemble based modelling approach has on human activity recognition.
First, integrating more than one base learner into a recognition system indeed has the positive effect we hypothesised at the outset of our work. 
In line with related work (cf.\ Sec.\ \ref{sec:related:ensembles}), it can be seen that strong learners --LSTM models in our case-- can indeed be integrated into very effective ensembles that outperform single, that is, individual models.
Recognition results are consistent across all three tasks, which is indicated by the relatively small standard deviation (Tab.\ \ref{tab:results}), which essentially demonstrates the effectiveness of our novel training method that generates models that focus well on different aspects of the (challenging) sample data.
Recognition results improve for both loss functions when increasing the number of base learners.
Practically for all three tasks the classification performance plateaus for $M = 10$ base learners (no statistically significant improvements when increasing to $M = 20$ (with the exception of Opportunity where doubling the number of base learners to $M = 20$ [F1 loss] leads to improvements that are just significant).

%
{\color{black} 

The classification performance can be further increased for Opportunity when injecting an additional  source of variation.
For this  dataset --which shows a strongly biased class distributions-- the combination of learners trained using different loss functions (CE and F1) further increases the diversity covered by the resulting ensemble, which pushes the recognition performance in a (strongly) significant manner ('Ensemble (M = $20$, 10 each) -- CE(10)+F1(10)' in Tab.\ \ref{tab:results}).
PAMAP2 does not benefit from diversity in loss function, which is attributed to the more homogeneous underlying class distribution and thus the lesser need for diversification of the modelling approach. Note that our LSTM based approach still substantially outperforms the state-of-the-art for all three tasks (see below) and that adding further diversity does not harm those recognition systems that seemingly have plateaued in their classification performance (such as PAMAP2), which is of  value for practical applications.}
} 

\begin{table}[t]
	\centering
	\tbl{ {\color{black} Results [F1 scores; standard deviation] for sample-wise activity recognition using variants of Ensembles of
LSTM learners. See text for description.\vspace*{1em}} }
	{
	\begin{tabular}{|l||c|c|c|}
		\hline
		Modelling Method			 				& Opportunity 	& PAMAP2 	& Skoda   		\\
		(\# learners M; loss function [F1, CE, both]) & & & \\
		\hline
		\hline
		\textbf{Single} (M = $1$) -- CE(1)	& $0.674 \pm 0.021$  	& $0.802 \pm 0.045$	& $0.918\pm 0.007$	\\
		\textbf{Single} (M = $1$) -- F1(1)	& $0.654 \pm 0.027$ 	& $0.776\pm0.050$ 	& $0.907\pm0.030$ 		\\
		\hline
		\textbf{Ensemble} (M = $10$) -- CE(10)	& $0.716 \pm 0.011$  	& $0.848\pm 0.021$	& $0.925\pm0.002$	\\
		\textbf{Ensemble} (M = $10$) -- F1(10)	& $0.706 \pm 0.013$ 	& $0.810\pm0.033$  	& $0.921\pm0.003$  	\\
		\hline
		\textbf{Ensemble} (M = $20$) -- CE(20)	& $0.720 \pm 0.011$  	& $0.852\pm 0.026$	& ${0.926}\pm0.001$	\\
		\textbf{Ensemble} (M = $20$) -- F1(20)	& $0.714 \pm 0.012$ 	& $0.811\pm 0.032$  	& $0.921\pm 0.002$  	\\
		\hline
		\textbf{Ensemble}  (M = $20$; $10$ each) -- CE(10)+F1(10)	& ${0.726} \pm 0.008$ 	& ${0.854}\pm0.026$ 	& $0.924\pm0.002$		\\
		\hline
	\end{tabular}}
	\label{tab:results}
\end{table}
%

The results of the experiments with different loss functions during model training are interesting in itself.
Cross entropy loss functions are widely used in the literature. 
However, when evaluating the overall recognition accuracy using $F1$ scores --as it is ideal for imbalanced data scenarios-- it seems more intuitive to use the alternative $F1$ loss function for model optimisation.
Interestingly, we did not see any positive effect on individual learners when changing the loss function.
One of the major reason is that the $F1$ loss is a global measure \cite{Fmeasure}, while our models are updated based on mini-batches, which may not precisely approximate the "true" gradient towards an optimal F1 score over the whole data set.
Given the features of both loss functions, we expected these two types of base learners may behave differently in certain scenarios, and are likely to be less correlated. 
The experimental results confirm our expectation, and we can see the Ensemble (CE+F1)
leads to improved performance for the most challenging recognition task.
Arguably, this further justifies our motivation for using Ensembles of LSTMs.
By mixing learners that were trained with different loss functions we further increase the variability as it is of importance for any classifier Ensemble approach \cite{Kittler1998-OCC}.

{\color{black}
Fig.\ \ref{fig:results-class} provides insights into which activities benefit most from our approach. 
In line with our original hypothesis the baseline classifiers struggle the most with those classes that are rather diverse:
\begin{description}
\item[Opportunity] Given the challenging nature of the dataset and the variability of its activity instances (cf.\ Fig.\ \ref{fig:durations} for a visualisation of the substantial variability in duration alone), nearly all target activities benefit from our  approach that utilises a combination of base learners that are based on the two different loss functions (i.e., CE(10)+F1(10)) achieves the best performance. 

\item[PAMAP2] For this rather homogeneous dataset, performance gains are more moderate (still significant) and are largely centred on 'sitting', 'stairs', and 'vacuuming'. Given that humans typically engage in a range of parallel activities while performing either of these three activities 
this is a result that supports our hypothesis that modelling needs to focus on diversity. 

\item[Skoda] In line with the previous argumentation the Skoda task benefits the most from our approach for the opening and closing doors activities. 
\end{description}
Overall, our method --that is the Ensemble with combined loss functions 'CE(10)+F1(10)'-- produces the largest performance gains for challenging classes and almost never has a detrimental effect on any of the activity classes, which is very encouraging for real-world scenarios.

For completeness Fig.\ \ref{fig:results-confusion} shows confusion matrices for all three recognition tasks (each averaged over the $30$ trials) and the best performing model configuration ('Ensemble (M=$20$) -- CE(10)+F1(10)'; cf.\ Tab.\ \ref{tab:results}). 
Not surprisingly, the largest confusion is caused by the NULL class (if there is one). Also, classes with very small sample numbers do not benefit as much due to the fact that our learners simply will not be able to capture as much variability as for larger sample sets.
}








To contextualise the achieved results we ran reference experiments that allow us to compare our approach to the state-of-the-art. 
{\color{black}Note that, according to the literature, the best results on all three datasets are currently achieved using (variants of) deep learning methods. Consequently, our comparison is focused on these approaches that effectively serve as benchmarks at the time of writing.}
These reference experiments are based on implementations as described in the particular original papers thereby in part making use of the source code provided by the authors and replicating the model configurations used there \cite{Hochreiter1997,Hammerla2016,Ordonez2016}.
Tab.\ \ref{tab:reference} summarises the results illustrating the substantial improvement in recognition performance for both Opportunity and PAMAP2.

When analysing our results in light of the class distributions of the particular datasets (Fig.\ \ref{fig:datadistr}) and the general quality of the sensor readings --as discussed throughout this paper these were the main motivation for the developing the Ensemble approach-- it becomes clear that our method is well suited for real-world, challenging activity recognition tasks.
Specifically for those scenarios with imbalanced class distributions as they are typical for mobile application scenarios and with challenging data quality Ensembles of LSTM learners very effectively make use of said data resulting in real gains of recognition performance.
The Opportunity challenge is widely known for being both realistic and challenging in the aforementioned sense and observing substantial improvements through employing our method is very encouraging.
The same holds for the PAMAP2 task, though not at the same magnitude, which is as expected considering the less biased dataset resulting from more controlled recording settings.
Somewhat surprisingly no substantial --yet still significant-- performance boost can be gained for the Skoda task, which we believe is attributed to the fact that only data for one worker is considered 
{\color{black}and the task thus does not include challenging test data from different environments for robustness evaluation. 
In this case, our method does not benefit significantly from its generalisation capability.
For this simple task,} 
the baseline models already achieve high specialisation.
Evidence for this assumption is given by the very high baseline results ($\bar{F_1} >0.9$).
However, even though our method does not increase recognition accuracy here, it also does not cause any harm, which is very encouraging for real world applications as neither any assumptions nor prior knowledge about the underlying class distribution are required for using our approach.

\begin{table}[t]
	\centering
	\tbl{{\color{black}Comparison of recognition results achieved using our LSTM Ensemble (CE(10)+F1(10)) vs baselines using the state-of-the-art for sample-wise activity recognition (F1 scores averaged over $30$ trials each).}}{
	\begin{tabular}{|l||c|c|c|}
		\hline
		Modelling Method			 				& Opportunity 	& PAMAP2 	& Skoda   		\\
		\hline
		LSTM baseline (acc.\ to \cite{Hochreiter1997})		& $0.659$ 	& $0.756$ 	& $0.904$		\\
		DeepConvLSTM (acc.\ to \cite{Ordonez2016})		& $0.672$ 	& $0.748$		& $0.912$    	\\ 

		LSTM-S (acc.\ to \cite{Hammerla2016})			& $0.684$ 	& $0.838$ 	& $0.921$		\\
		LSTM Ensemble (this work)					&  {\color{black}0.726  $\pm 0.008$ } &{\color{black}0.854 $\pm 0.026$}&  {\color{black}0.924$\pm 0.002$}    \\ 
		
		\hline
	\end{tabular}}
	\label{tab:reference}
\end{table}

%% file: discussion.tex
\section{Discussion}
\label{sec:discussion}
The main motivation for our work is to develop innovative modelling techniques for robust and reliable human activity recognition that can be deployed in real-world scenarios.
Very much in the general tradition of Ubicomp and its original vision we strive to have a positive impact on people's lives, and activity recognition in ubiquitous computing forms the basis for many research endeavours that follow that vision.
As such, the recognition capabilities of state-of-the-art HAR methods for challenging, that is, real-world scenario is still leaving substantial room for improvement. 
For example, even with the latest generation of deep learning modelling techniques, benchmark results for the Opportunity challenge --arguably one of the most challenging yet most realistic HAR datasets-- still suggests that substantial further research is needed on the way towards fulfilling the Ubicomp vision.

In this paper we developed an activity recognition framework that targets challenging, real-life scenarios, specifically addressing noisy and imbalanced datasets as they are typical for mobile applications of human activity recognition.
Our framework is based on the integration of individual deep Long Short Term Memory (LSTM) networks into Ensemble classifiers.
{\color{black}We chose deep learning in general and (unmodified) LSTM models in particular as the starting point for our work because recent related work employing LSTM recognisers has demonstrated their excellent recognition capabilities including superior classification performance for many applications in the wider wearable and ubiquitous computing domain.}
LSTMs represent variants of recurrent neural networks with principally infinite memory.
In line with recent related work we argue that such sequential models are well suited for the analysis of sensor data streams.
Through combining multiple, diverse base LSTM learners into Ensembles we are able to significantly increase the robustness of the resulting recognition system, which is illustrated by substantial improvements of the recognition accuracy as validated on three standard benchmark datasets (Opportunity, PAMAP2, Skoda).
The core technical contribution of our work can be summarised as follows:
\begin{enumerate}
\item To the best of our knowledge this is the first time that deep LSTM models are integrated into an Ensemble based modelling approach for human activity recognition using wearables.
With this Ensemble framework we are able to significantly improve recognition accuracies on standard benchmark datasets.
{\color{black}These datasets represent challenging, real-world HAR scenarios and significantly improving over the state-of-the-art demonstrates the practical value of our proposed method.}

\item The basis for our Ensemble framework is a modified training procedure for deriving diverse sets of LSTM based activity recognisers that specifically target problematic data situations in real-life scenarios.
{\color{black}The anatomy of the underlying base LSTM learners remains unchanged (w.r.t.\ how they have been introduced in the literature), which, together with its implementation using standard, open-source toolkits, allows for straightforward replication of our approach (and experiments) and system integration.}

\item We employ our Ensembles of deep LSTM networks for improved, sample-wise activity recognition, which has substantial potential for future applications of HAR, such as, activity prediction for time-critical inference.
\end{enumerate}

{\color{black}Our extensive experimental evaluation demonstrates the potential the presented approach has. For all three analysed standard benchmark datasets the presented approach outperforms the state-of-the-art in a statistically significant manner. Specifically for challenging HAR scenarios --exemplarily explored using the Opportunity dataset-- substantial gains of classification accuracy can be achieved. Our method effectively tackles typical problems of real-world deployments of wearable HAR systems namely imbalanced class distribution often with substantially over-represented NULL / background class, as well as noisy if not faulty sensor readings. We have argued that in real-world deployment it is typically not straightforward to filter out unwanted samples, such as background or NULL data, and thus the increased robustness provided by our approach is especially encouraging.
Arguably, our method has the greatest potential for those challenging datasets. In addition the results of our experimental evaluation demonstrate that the method does barely lead to detrimental results even for the cases where single models already achieve very high classification accuracies (some activities in PAMAP2 and Skoda). As such our experiments demonstrate a generalisable applicability.
}

{\color{black}
The focus of this paper was on the development of the basic methodology. As such we have not yet addressed implementation issues as they would be of relevance for field deployments or products even. However, it has recently been shown that even complex deep neural networks can effectively be used for inference on substantially resource constrained platforms such as smartwatches \cite{Bhattacharya:wd}. In order to be able to implement complex deep neural networks on such devices sophisticated optimisation techniques are necessary. The majority of these correspond to pruning efforts with regards to the effective parameter space. Our Ensembles certainly inflate this parameter space, yet the general principle of pruning remains accessible as we did not change the general nature of the models (the core of our contribution lies in the modified training procedure) -- no dependencies have been introduced that would prevent pruning of the parameter space. As such we are confident that our models can be implemented on wearable devices for interactive HAR scenarios. Note that offline analysis scenarios --as they are most common in the Ubicomp domain, for example, for after the fact analysis in health and wellbeing domains-- are not affected at all by such concerns and our method is directly usable.
Importantly our models operate on the basis of sample-wise prediction. This is a principle advantages for future real-time applications (e.g., fall detection). 
}

%% file: appendix.tex
\appendix
\section*{APPENDIX}
\setcounter{section}{0}
\section{Reducing Loss in LSTM Ensembles}
\label{app:loss}
{\color{black} 

Our recognition approach integrates a number of vanilla LSTM models as base learners into an Ensemble framework. By means of our experimental evaluation we have demonstrated how this approach leads to significantly improved robustness and thus classification performance.
In what follows we provide an exemplary and more formal exploration on why LSTM base learners lead to improved classification performance.

As a sequential model, LSTM takes a current signal $\mathbf{x}_t$, and previous states (i.e., hidden state $\mathbf{h}_{t-1}$ and cell state $\mathbf{c}_{t-1})$ as inputs and performs sample-wise prediction (Eqs.\ (\ref{eq:lstmFF}) -- (\ref{eq:softmax}) in Sec.\ \ref{sec:related:LSTM}).
For the sake of clarity but without loss of generality, here we argue based on the example of reduction of cross entropy loss (referred to as CE), arguably the most popular loss function in the field.
Specifically, CE corresponds to the negative logarithmic loss of the predicted probability for target class $k$ and thus --inversely proportionally-- to the predicted class probability.
For example, with class probability $p_{tk}$ for target class $k$ at time $t$, the corresponding CE $L^{CE}_t$ is:
\begin{equation}
	\label{eq:log_loss}
	L^{CE}_t=  -\ln p_{tk}. 
\end{equation}
Let $\mathbf{z}_t \in \mathbb{R}^K$ be the binary vector indicating correct class label $k$ such that $z_{tk}=1 \land \{z_{tj}=0\}_{j\neq k}$.
$L^{CE}_t$ can also be written in vector form as follows: 
\begin{equation}
	\label{eq:avgLossCE}
	L^{CE}_t=  -(\mathbf{z}_t)^T\ln \mathbf{p}_t. 
\end{equation}
Next we explicitly model the process of reducing LSTM loss --CE in this case-- for unseen test data. 


\subsection*{Average Loss of $M$ Individual Learners}

For test data with $t=1,2,..., N$ samples, the expected CE $L^m$ of the $m$-th model is given as follows:
\begin{equation}
	\label{eq:avgLossCE}
	L^m=  -\mathbb{E}_t(\mathbf{z}_t)^T\ln \mathbf{p}^m_t, 
\end{equation}
where $\mathbb{E}$ denotes expectation.
With this the average loss $L^{avg}$ for $M$ models can then be written as:
\begin{eqnarray} 
\label{eq:loss_avg}
	L^{avg} 	& = & - \frac{1}{M}\sum^M_{m=1} \mathbb{E}_t(\mathbf{z}_t)^T\ln \mathbf{p}_t^m\\
			& = & -\mathbb{E}_t \overline{\ln{p_{tk}}} \nonumber \\
			\text{with:} \quad	\overline{\ln{p_{tk}}}	& = & \frac{1}{M}\sum^M_{m=1}\ln p_{tk}^m.
\end{eqnarray}

\subsection*{Loss when fusing $M$ models}
For a fused model, i.e., an Ensemble that has been created through score level fusion of multiple individual learners, the corresponding CE can be formulated in a similar way.
At time step $t$ the CE $L^{fusion}_t$ of the Ensemble is defined as follows:
\begin{equation}
	\label{eq:ensembleLoss}
	L^{fusion}_t = -(\mathbf{z}_t)^T \ln \mathbf{p}_t^{fusion}
\end{equation}
where $\mathbf{p}_t^{fusion}$ denotes the score level fusion, as defined in Eq. (\ref{eq:fusion}).
The average CE $L^{fusion}$ for the $t=1,2,...,N$ test samples can then be written as:
\begin{eqnarray} 
\label{eq:lossFusion}
	L^{fusion} 	& = & -\mathbb{E}_t(\mathbf{z}_t)^T \ln \mathbf{p}_t^{fusion}  \\
			& = & -\mathbb{E}_t(\mathbf{z}_t)^T \ln  (\frac{1}{M}\sum_{m=1}^M{\mathbf{p}^{m}_t}) \\
			& = & -\mathbb{E}_t\ln \overline{p_{tk}} \\
			\text{with:} \quad \overline{p_{tk}} & = &\frac{1}{M}\sum^M_{m=1}p_{tk}^m.
\end{eqnarray}
Let $\Delta = L^{avg} - L^{fusion}$ denote the difference between average CE of individual LSTM learners and the CE of the Ensemble for the test data, then we have:

\begin{eqnarray} 
\label{eq:loss_diff}
	\Delta  	& = & -\mathbb{E}_t\{ \overline{\ln{p_{tk}}}  -   \ln \overline{p_{tk}}         \} \\
		 	& = & -\mathbb{E}_t\{ \frac{1}{M}\sum^M_{m=1}\ln p_{tk}^m -   \ln \frac{1}{M}\sum^M_{m=1}p_{tk}^m        \} \\
			& = & -\mathbb{E}_t\{ \frac{1}{M}\ln\prod^M_{m=1} p_{tk}^m -   \ln \frac{1}{M}\sum^M_{m=1}p_{tk}^m        \} \\
			& = &  -\mathbb{E}_t \{ \ln  \sqrt[M]{\prod^M_{m=1} p_{tk}^m}  -   \ln \frac{1}{M}\sum^M_{m=1}p_{tk}^m   \}\\
			& = &  -\mathbb{E}_t \{ \ln  \frac{\sqrt[M]{\prod^M_{m=1} p_{tk}^m}}  {\frac{1}{M}\sum^M_{m=1}p_{tk}^m}   \}.
\end{eqnarray}
Since $ \sqrt[M]{\prod^M_{m=1} p_{tk}^m} \le {\frac{1}{M}\sum^M_{m=1}p_{tk}^m}$ holds for $\{p_{tk}^m \ge0 \}_{m=1}^M$ (inequality of arithmetic and geometric means \cite{AMGM2004}), we have shown that the average model loss for the Ensemble is always less or equal to the average loss of individual learners, which is beneficial for the targeted application scenario:
\begin{equation} 
\label{eq:fusion_good_CE}
L^{avg} \ge L^{fusion}.
\end{equation}
Note that $L^{avg} = L^{fusion}$ holds if $p_{tk}^1=p_{tk}^2=...=p_{tk}^M$, i.e., when fusing $M$ identical LSTM learners. 
In general, the Ensemble performance benefits (with improved prediction probabilities for the true classes) by fusing $M$ diverse, i.e., less correlated models.

}